\documentclass[a4paper, oneside]{cspgthesis}
\usepackage{geometry}
\usepackage[utf8]{inputenc}
\usepackage[T1]{fontenc}
\usepackage{graphicx}
\usepackage{algorithm}
\usepackage{algpseudocode}
\usepackage{amsmath}
\usepackage{float}
\usepackage{pdfpages}

\usepackage{hyperref} 
\usepackage{xurl}

\usepackage{bbm}

\definecolor{mcts_green}{RGB}{56, 118, 29}
\definecolor{mcts_red}{RGB}{204, 0, 0}
\definecolor{mcts_blue}{RGB}{17, 85, 204}

\usepackage{booktabs}
\usepackage{multirow} 
\usepackage[section]{placeins}
\newcommand{\na}{\textendash}
\newcommand{\pct}{(\%)} 

\thesistype{Bachelor's Thesis} 
\title{Creating Steering Vectors via Input Optimisation}

\author{Batuhan Akar}
\email{baakar@ethz.ch}

\institute{Secure and Private AI Lab \\[2pt]
ETH Zürich}

\supervisors{Lukas Fluri, Prof.\ Dr.\ Florian Tramèr}

\date{August 25, 2026}

\begin{document}

\frontmatter 
\maketitle

\cleardoublepage

\cleardoublepage
  \thispagestyle{empty}
  \vspace*{0.28\textheight}
  \begin{flushright}{\itshape\LARGE To Dennis}\end{flushright}
  \vspace*{\fill}
\cleardoublepage

\begin{acknowledgements}
I'm grateful to Lukas Fluri, my supervisor, for guiding me throughout this thesis. Without his support, this project would not have been possible. I'm also grateful to the whole SPY Lab team, as their amazing environment has helped me stay strongly motivated throughout the semester. Last but not least, I am thankful to my friends and family, especially my brother Dennis, and my parents Tolga and Banu, for always being there for me.

Thank you.
\end{acknowledgements}



\tableofcontents

\mainmatter 


\begin{abstract}
Activation steering influences a language model's behaviour by adding a steering vector to its residual stream during generation, but it requires white-box access to modify the hidden states at every generation step. We introduce \textbf{ActGCG}, a modification of Greedy Coordinate Gradient which optimises a suffix to match the target activations obtained by adding a steering vector to the residual stream. We further introduce universal suffixes trained across many prompts at once, a model-translation step that rewrites the suffix into human-readable text midway through optimisation, and a soft-suffix variant to remove the discrete token space constraint. We study two behaviours: steering a model to answer in German, and anti-refusal steering for jailbreaking. For German, a discrete suffix reproduces the steering vector's behaviour almost perfectly and transfers well across prompts and models. For anti-refusal, the equivalence is weaker, succeeding under a phrase-avoidance metric but only partially under the stricter HarmBench classifier. Our results show that a steering vector's behaviour is reachable from the input, with no access to model internals at generation time, up to a loss that depends on the concept being steered.

\end{abstract}

\chapter{Introduction}%
\label{chap:intro}

The behaviour of large language models can be influenced by directly editing their internal activations. A common method of achieving this is to use \emph{activation steering}, where one computes a \emph{steering vector} to induce a target behaviour and adds the vector to the residual stream during generation to push the model towards it. One can build such a vector by averaging the difference in activations
between two datasets with prompts that do and do not exhibit the behaviour \citep{rimsky2024caa}. Steering vectors are useful as they're cheap to produce and interpretable. However, since they are an intervention in the activation space, one requires full access to the model's weights as well as the ability to modify hidden states at every generation step. This is not feasible when a model can only be reached through a chat interface.

At the same time, model behaviour can be changed automatically from the input alone. Greedy Coordinate Gradient (GCG) \citep{zou2023gcg} is a suffix finding method to cause a model to yield an affirmative string such as ``Sure, here is how to\ldots'', also called the target string. The adversarial suffix is optimised by only changing the suffix tokens themselves to maximise the chance of obtaining the target as output, without directly modifying internal activations.

We look for an equivalence relationship between suffixes and steering vectors by combining these two methods, and investigating whether the effect of a steering vector can be reproduced by an input suffix. To this end, we modify the loss function of GCG to reach a new algorithm, which we call \textbf{ActGCG}. Instead of targeting an output string, ActGCG targets the \emph{activations} that
a steering vector would have induced. The vector is first given on a set of filler tokens to obtain the target activations. The algorithm then searches for a discrete suffix whose activations are on those same tokens to match the target, minimising the squared difference summed over positions and over several layers. If such a suffix exists, it means that the steering vector's
behaviour was reachable from the input, with no access to the model's internals at generation time. 

We study two behaviours. The first task is to steer a model to answer in \textbf{German}, using a contrastive German/English steering vector built from the Dolly instruction dataset \citep{databricks2023dolly}. The second is \textbf{anti-refusal} steering, where the vector is negated so that matching the target pushes the model \emph{away} from refusing a harmful request, and we measure success with the HarmBench
jailbreak classifier \citep{mazeika2024harmbench} as well as checking if a model avoids a set of refusal phrases. For German, the discrete suffix reproduces the steering vector's behaviour almost perfectly. For the anti-refusal direction, we achieve a high success rate for phrase avoidance, but lower success rate for jailbreaking.

Beyond the core method, we make three further contributions. First, we introduce a \emph{soft-suffix} variant that replaces the discrete tokens with continuous embeddings optimised directly by gradient descent. We use this as an upper bound on how well the activation-matching objective can be met, as it is unconstrained by the discrete token space. Second, we move from single prompts to \emph{universal} suffixes: a single string trained across multiple prompts at once, which we then use to study transfer across new prompts of the same model, and across different models. Third, we add a \emph{model-translation} step to the universal setting, where in the middle of the optimisation steps the unreadable suffix is rewritten into fluent, human-readable text using another model. We then continue the optimisation as before.

The rest of the thesis is organised as follows. Chapter~\ref{chap:background} reviews the linear representation hypothesis, steering vectors, and the GCG algorithm. Chapter~\ref{sec:methodology} introduces the ActGCG method and the experimental setup. Chapter~\ref{sec:experiments} presents our experiments on the German and anti-refusal directions, covering per-prompt, universal, model-translation, and soft
suffixes, as well as transferring across models. Chapters~\ref{chap:discussion} and~\ref{chap:conclusion} discuss the findings and conclude.

\chapter{Background}%
\label{chap:background}

We first discuss the internal mechanisms of a Large Language Model, explain the role of steering vectors in this context, and elaborate on the GCG algorithm.

\section{Layers and Tokens}

Large Language Models are built up of stacks of layers between an input and an output. These layers allow the model to represent increasingly complex functions. Each layer consists of a Multi-Layer Perceptron (MLP) and multiple attention heads. The MLP processes information in each token position, whereas attention moves information between token positions. 

A token is a unit of text that the model can process, represented by a vector of numbers. We have two notions of a token. A \emph{discrete} token corresponds to a word of the model's fixed vocabulary. In contrast, a \emph{soft} token is an arbitrary vector in embedding space that does not necessarily correspond to a word in a vocabulary. Hence, allowing soft tokens gives an optimiser access to the entire continuous embedding space rather than the discrete vocabulary.

\section{Activations}

For each layer and each token position, the transformer~\citep{vaswani2017attention} of a Large Language Model maintains a vector that is updated by the attention and MLP sublayers. Attention combines vectors from earlier positions, and the MLP applies a non-linear transformation per position. The transformer adds each sublayer's output back onto its input through a
\emph{residual connection} \citep{vaswani2017attention}. More formally, following the residual-stream view of \citet{elhage2021framework} and ignoring layer normalisation for clarity, we let $h_i^{(\ell)}$ be the vector at token position $i$ after layer $\ell$. We first have attention write into the stream:
\begin{equation}
  \tilde{h}_i^{(\ell)} = h_i^{(\ell)} + \mathrm{attn}^{(\ell)}\!\big(h_{\le i}^{(\ell)}\big),
  \label{eq:attn}
\end{equation}
Then, we let the MLP read that post-attention value and write into it:
\begin{equation}
  h_i^{(\ell+1)} = \tilde{h}_i^{(\ell)} + \mathrm{mlp}^{(\ell)}\!\big(\tilde{h}_i^{(\ell)}\big),
  \label{eq:residual}
\end{equation}
Therefore, each sublayer writes into a running sum. This vector is variously called the hidden state, the activation vector, or the residual stream, and it lives in a high-dimensional \emph{activation space}. The residual form of Equation~\ref{eq:residual} is what makes activation steering possible, as an intervention added to $h_i^{(\ell)}$ is carried forward to every subsequent layer.

The \emph{linear representation hypothesis} claims that human-interpretable concepts are represented along linear subspaces\citep{mikolov2013b, pennington2014, bolukbasi2016, elhage2022, park2023, nanda2023}, i.e. as directions in the activation space. We adopt this assumption in our work. For example, if a concept corresponds to a direction $v$, we assume that moving an activation along $v$ strengthens the concept. The difference between the mean activations of examples that do and do not exhibit the concept is an estimate of $v$, which we explore as steering vectors.

\section{Prompt structures with chat templates}

Instruction-tuned models wrap user text in a \emph{chat template}. A template uses additional tokens to determine where the user begins and ends, and where the assistant turn starts. The number of chat template tokens differs from model to model, which can change the optimal token position to extract steering vectors. For our method, since we read the activations at fixed positions between the user prompt and the response, the chat template determines the token positions to be extracted. We refer to this as \emph{filler} positions as given in Chapter~\ref{sec:methodology}.

\section{Steering Vectors}

Under the linear representation hypothesis, we can influence a model's output by adding a vector to the residual stream at a chosen layer and token position. Concretely, given a steering vector $v$ for a target concept and a strength $\alpha$, we replace the activation during generation by
\begin{equation}
  h_i^{(\ell)} \;\leftarrow\; h_i^{(\ell)} + \alpha\, v,
  \label{eq:steer}
\end{equation}
which pushes the model towards the concept for $\alpha>0$ and away from it for $\alpha<0$.
Steering vectors can nudge models along behaviours of varying complexity: the output language, refusal, sycophancy (telling the user what they want to hear), and so on. They are computationally easy to build and interpretable by humans.

\begin{figure}[H]                       
    \centering
    \includegraphics[width=0.6\textwidth]{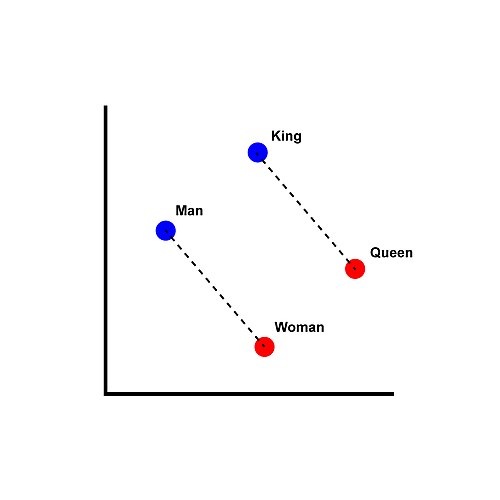}
    \caption{Semantic relationships as directions in an embedding space.\protect\footnotemark}
    \label{fig:word-vectors}
  \end{figure}

The underlying intuition predates language models. In classical word embeddings, semantic
relationships appear as consistent directions, so that vector arithmetic such as
$\text{king}-\text{man}+\text{woman}\approx\text{queen}$ recovers analogies
(Figure~\ref{fig:word-vectors}). Steering vectors carry this intuition into the residual stream of a transformer.

  \footnotetext{\url{https://commons.wikimedia.org/wiki/File:Word_vector_illustration.jpg}}

\section{Steering Vector Algorithms}

Several methods exist for constructing the direction $v$; they differ mainly in how many contrast examples they use, at which layer and position they read activations, and whether they apply a further linear transformation. We summarise the most common ones.

\emph{Activation Addition (ActAdd)}~\citep{turner2023actadd} takes a single contrast pair, runs both prompts through the model, records the activations at one layer, and subtracts them to obtain the steering vector, which is then added at one token position during generation.

\emph{Contrastive Activation Addition (CAA)}~\citep{rimsky2024caa} generalises to many contrast pairs. For each pair it computes the positive and negative activations at a single layer and last token position and takes the mean of their differences across all pairs, adding
the resulting vector at every token position during generation. CAA is evaluated on both multiple-choice and open-ended prompts. Our own steering vectors follow this difference-of-means
recipe (Chapter~\ref{sec:methodology}).

\emph{Inference-Time Intervention (ITI)}~\citep{li2023iti} considers the attention-head space rather than the full residual stream. Given a dataset of true/false question--answer pairs,
it records the activation of each attention head, trains a linear probe per head to separate
true from false, and keeps the top-$K$ most predictive heads. For each such head it takes the
mean-difference direction between true and false examples and shifts activations along the direction, targeting truthfulness specifically.

\emph{Representation Engineering (RepE)}~\citep{zou2023repe} collects contrast pairs where the contrast is only in the target concept, and reads activations (by default at the last token) for both. For each layer it forms the difference vectors and runs principal component analysis over them; the first principal component is taken as the reading direction for that concept at that layer, which can then be used either
as a classifier (by projection) or for control (by addition or subtraction).

Steering based on sparse autoencoders (SAEs), which impose a sparsity constraint so that only a few hidden units activate at once, is also possible but is harder to evaluate and has not shown
a clear advantage in comparable settings.\footnote{See the discussion in
\url{https://www.research-collection.ethz.ch/server/api/core/bitstreams/f4073b47-2050-4194-8540-a8a43eaf8ecb/content}.}

We adopt the simpler and well-supported difference-of-means (CAA-style) construction.

\section{GCG mechanism}

Greedy Coordinate Gradient (GCG)~\citep{zou2023gcg} optimises a suffix appended to a prompt such that the model begins its answer with a chosen affirmative \emph{target string}. For example, this can be a result starting with ``Sure, here is how to\ldots''. To obtain such a suffix, the algorithm works in the following manner: 

For a given prompt, our loss function is the cross entropy (i.e. the ``surprise'') of the target tokens given the prompt and the suffix, which we minimise. For a current suffix, we first take the embeddings of the tokens and find the gradients of the loss with respect to these embeddings. We compute the top-k promising tokens per position for a replacement. Then, for each batch, we choose a random token position in the suffix, and replace it with a uniformly sampled token from the top-k options. We receive a number of candidate suffixes, and choose the suffix of the batch that minimises the loss (and hence maximises the chance of reaching our target string). We repeat this T times.

\begin{figure}[htbp]
  \centering
  \includegraphics[width=0.9\textwidth]{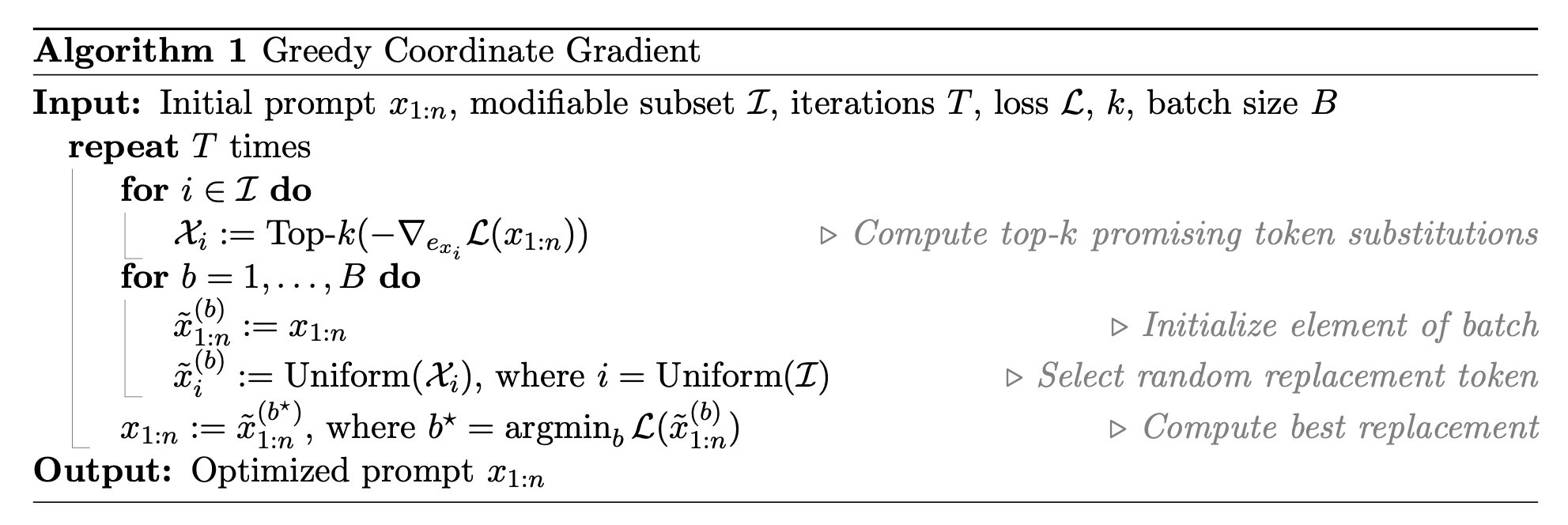}
  \caption{The Greedy Coordinate Gradient (GCG) algorithm, reproduced from \citet{zou2023gcg}.}
  \label{fig:gcg}
\end{figure}

Crucially, GCG searches over \emph{discrete} tokens. Figure~\ref{fig:gcg} states the algorithm formally.

\section{GCG modifications and alternatives}

There are numerous ways of modifying GCG, as well as alternatives to the GCG algorithm. We explore three such methods.

\emph{Steering-SSR}~\citep{winninger2025ssr} is the most similar prior method. It optimises a suffix so that the model's activations match a target where a steering vector is added, but using a cosine-similarity loss rather than a squared distance.\footnote{Eq.~\ref{eq:ssr-loss} from \texttt{SteeringSSR} in \url{https://github.com/Sckathach/subspace-rerouting}.} Writing $h_\ell$ for the suffix's readout activation and $v_\ell$ for the unit steering direction at layer $\ell$, its loss sums a per-layer squared cosine term over the chosen layers,
\begin{equation}
  \mathcal{L}_{\text{SSR}}(s) \;=\; \sum_{\ell \in \mathcal{L}}
        \big(\,{-}\alpha_\ell \;-\; \cos\!\big(h_\ell,\, v_\ell\big)\big)^2,
  \label{eq:ssr-loss}
\end{equation}
where the per-layer target $\alpha_\ell$ sets the desired cosine. We were unable to reproduce the reported jailbreak success rates, since re-judging the paper's outputs with our own classifier (see Chapter~\ref{sec:methodology}) gave substantially lower rates than reported, which we attribute to the choice of jailbreak judge.

\begin{figure}[htbp]
  \centering
  \includegraphics[width=0.65\textwidth]{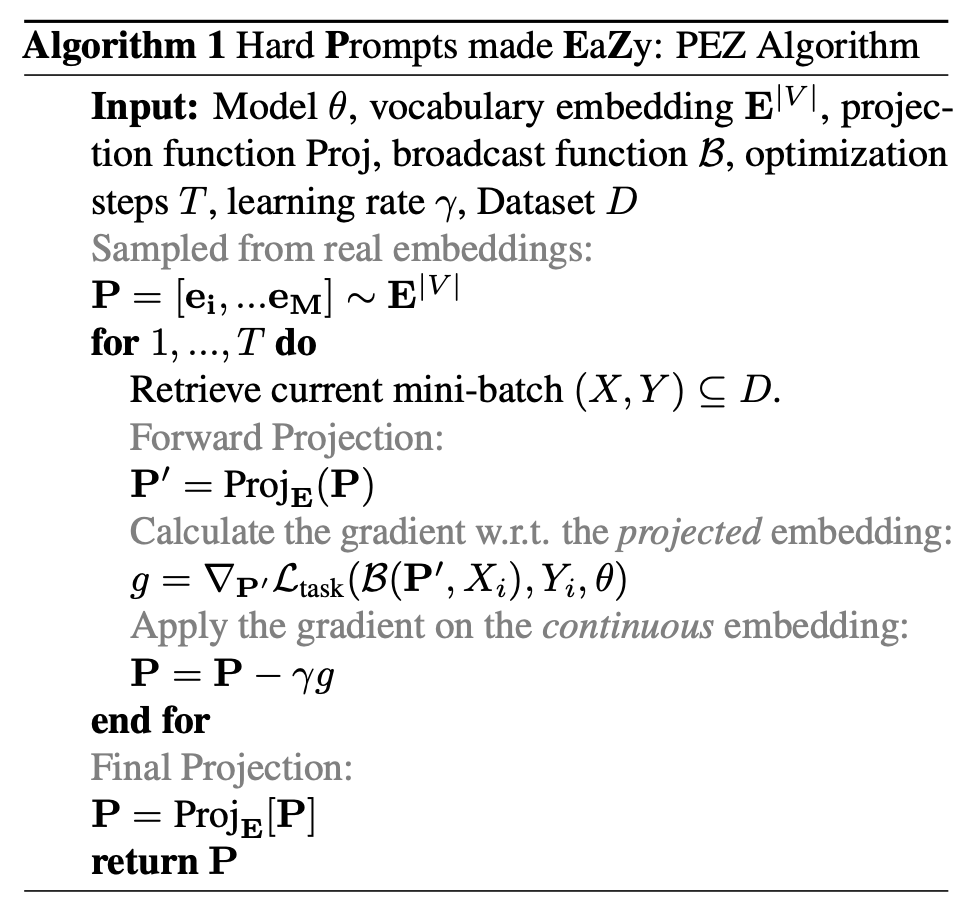}
  \caption{The PEZ gradient-based discrete optimisation scheme, reproduced from
  \citet{wen2023pez}}
  \label{fig:pez}
\end{figure}

\emph{PEZ (Hard Prompts Made Easy)}~\citep{wen2023pez} performs gradient-based discrete optimisation by combining soft and discrete updates. In each iteration it projects the current continuous embeddings onto their nearest discrete vocabulary tokens, computes the gradient at the
projected point, and applies that gradient to update the continuous iterate. A final projection yields the discrete output. Figure~\ref{fig:pez} states the algorithm formally.

Finally, we can remove the discrete token constraint altogether by considering \emph{soft suffixes}. We use this as a comparison method in Section~\ref{sec:models-judging}.

\chapter{Methodology}%
\label{sec:methodology}

We now introduce our method. We first describe how we construct the steering vectors that define each target behaviour, and the ActGCG objective that searches for a suffix reproducing the activations when a steering vector was added. Then, we follow with the universal and model-translation versions of the objective, our tasks and datasets, as well as the models, judges that were used together with the comparisons to other methods.

\section{Constructing Steering Vectors}
\label{sec:steering-vectors}  

We build a steering vector for each behaviour with a difference-of-means construction inspired by Contrastive Activation Addition~\citep{rimsky2024caa}. For a set of contrast pairs (such as German against English), we run both prompts through the model, and read the residual-stream activation at a chosen layer and token position. We average the difference across all pairs. Since our loss matches activations at several layers at once (Section~\ref{sec:ares}), we build one such vector per target layer and stack them together. For every model, we use four layers, with a preference towards the middle and later layers where steering vectors are commonly applied, as listed in Table~\ref{tab:layers}.

\begin{table}[htbp]
  \centering
  \begin{tabular}{lcc}
    \toprule
    Model & Total layers & Chosen layers \\
    \midrule
    Gemma-2-2B-it   & 26 & 10, 14, 18, 22 \\
    Qwen3-1.7B      & 28 & 10, 13, 16, 20 \\
    Llama-3.2-3B    & 28 & 12, 16, 20, 24 \\
    Qwen3-4B        & 36 & 14, 18, 22, 28 \\
    \bottomrule
  \end{tabular}
  \caption{The four layers at which each model's steering vectors are built and at which the ActGCG objective matches activations.}
  \label{tab:layers}
\end{table}

We choose a CAA style method because it is both simple to implement and effective relative to its alternatives. The best method to obtain steering vectors overall is not yet fully clear; for instance, \citet{im2025unified} reports that mean-difference vectors outperform other steering constructions, whereas \citet{pres2024reliable} argues that CAA is less optimal than what earlier work suggested. We proceed with it regardless, as we found no alternative that gives a clear improvement for our purposes.

\begin{figure}[htbp]
  \centering
  \includegraphics[width=0.6\textwidth]{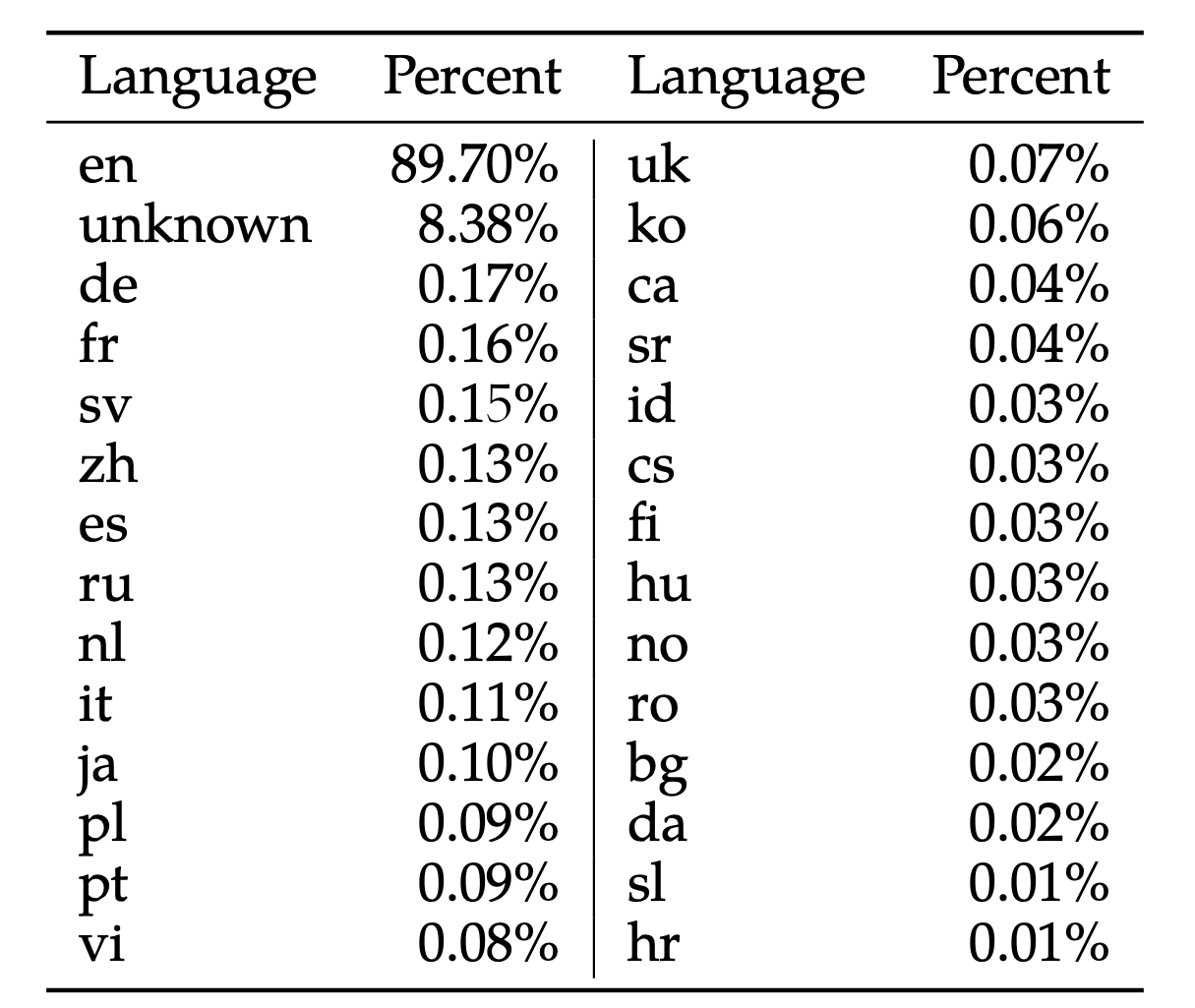}
  \caption{Language distribution in pretraining data with percentage >= 0.005\%. Most data is in English, meaning that Llama 2 will perform best for English-language use cases. The large unknown category is partially made up of programming code data. Reproduced from \citet{touvron2023llama2}.}
  \label{fig:llama_lan}
\end{figure}

For the language task, we first tried Turkish. However, we noticed that the level of knowledge for the language was quite limited. For example, Figure~\ref{fig:llama_lan} illustrates that Llama-2-7B has German as the second largest recognised language~\citep{touvron2023llama2}. As a result, we believed German might be the more optimal language choice. For the \textbf{German} vector, we use the Dolly instruction dataset~\citep{databricks2023dolly}. We pair each German (both question and response) prompt (positive) with its English counterpart (negative). We read the activations near the end of the response (position $-2$, or second to last token). The difference of means gives our ``answer in German'' direction. 

For the \textbf{refusal} vector we use the harmful and harmless instruction datasets of \citet{arditi2024refusal}, taking the harmful set as positive and the harmless set as negative. Here, unlike the German vector, there is no answer appended after the instruction, so we consider the last tokens before the response. The exact position changes depending on the model. We read at position $-1$ (last token of the prompt) for Gemma, Llama, and Qwen2.5~\citep{qwen2024qwen25}; but $-5$ for Qwen3, since its chat template also inserts an empty reasoning scaffold (\texttt{<think></think>}). Since the resulting vector points \emph{towards} refusal, to steer against refusal (the jailbreak direction) we negate it, which corresponds to using a negative strength $\alpha$.

Our vector extraction was implemented with the NNSight library~\citep{fiotto-kaufman2024nnsight}, which made our efforts simpler in tracing activations at specific layers and token positions. We do not claim our chosen layers and token extraction positions to be optimal (due to the amount of computational resources it would require). However, we chose our token and layer extraction positions after checking the success rates of multiple different combinations.

\section{The ActGCG Objective}
\label{sec:ares}

Our method, which we call \textbf{ActGCG}, modifies the loss of a GCG implementation (from the NanoGCG library~\citep{grayswan2024nanogcg}) while keeping its discrete search unchanged. Instead of a cross-entropy towards a target string,
we match the activations that a steering vector would have produced.

Concretely, let $x$ be a prompt and let the suffix be the tokens that follow it. We call the chat-template tokens that sit \emph{after} the user prompt and \emph{before} the generated response the \emph{filler} tokens, and read activations at those positions. We first build our target. For each target layer $\ell$ we run the plain prompt through the model, add
$\alpha\,v_\ell$ to the residual stream at every filler position, and read the resulting activations $t_{\ell,p}$ at filler position $p$. We then search for a suffix whose own forward pass reproduces those activations. Writing $a_{\ell,p}(x \mathbin{\Vert} s)$ for the activation at
layer $\ell$, filler position $p$, of the prompt with suffix $s$ appended, the loss is the
squared error summed over filler positions and over four layers:
\begin{equation}
    \mathcal{L}(s) \;=\; \sum_{\ell \in \mathcal{L}} \; \sum_{p \in \mathcal{F}} 
          \big\lVert\, a_{\ell,p}(x \mathbin{\Vert} s) \;-\; t_{\ell,p} \,\big\rVert_2^{\,2},
    \label{eq:ares-loss}
  \end{equation}
where $\mathcal{L}$ is the set of four layers and $\mathcal{F}$ the filler positions. Minimising
Equation~\ref{eq:ares-loss} with the ordinary GCG search lets us reproduce the steering vector's effect using a discrete suffix, without activation edits during generation.

We match at the filler positions rather than at the generated response since the response tokens change from run to run and differ across prompts, so the variance is quite high. The filler tokens,
by contrast, are a \emph{fixed} set of chat-template tokens that is present for every prompt and directly before generation. Reading the filler tokens provides a prompt-independent set of positions, and since there are several of them the steering vector signal is stronger from multiple positions at once rather than from a single token.

The variable $\alpha$ scales the strength of the steering vector. We sweep across different $\alpha$ for both the German and the anti-refusal directions. Unless noted otherwise, we optimise a suffix of $20$ tokens initialised to ``\texttt{x x \ldots\ x}'', for $250$ steps with search width $512$, top-$k=256$, replacing a single token per candidate, and a fixed seed for reproducibility.

\section{Universal Suffixes and Model Translation}
\label{sec:universal-mt} 

Our original objective is optimised on a single suffix for a single prompt. We extend our method in two ways to create suffixes that generalise across different prompts and models.

First, we consider a \textbf{universal suffix}. To obtain a single suffix that steers multiple prompts at once, we optimise one shared token string whose loss and gradient are \emph{summed} over a
training set of prompts. At each step, a token swap depends on the sum of the per-prompt gradients. We train on $N$ prompts, and evaluate on the training set as well as a set of new prompts to measure generalisation. To check model transfer, we evaluate the universal suffix on the same training prompt set across different models.

Second, a GCG suffix is typically unreadable to humans. To make it interpretable without losing its effect, we adopt the prompt translation idea from \citet{li2024translation} and insert a \textbf{model-translation} step into the middle of the optimisation steps. We run the universal search for the first half of the steps, and then make a single call to a larger \emph{proposer} model asking it to rewrite the current suffix into a coherent version of the same length. Afterwards, we resume the search from the rewritten suffix for the remaining steps. We also compare the mid-optimisation step against a variant that translates only once at the very end.

\section{Experimental Setup}

We test ActGCG across two concepts, German and refusal.\footnote{Code available at
  \url{https://github.com/ethz-spylab/s26-gcg-steering}.} We first optimise one suffix for each prompt to match the steering vector target activations at the filler tokens (Equation~\ref{eq:ares-loss}). We then test our method on universal suffixes that work across multiple prompts, as well as across multiple models. Finally, we apply \emph{model translation} to the universal suffix, having a larger model rewrite the optimised suffix into readable text midway through the optimisation (Section~\ref{sec:universal-mt}).

\paragraph{Models.} Our primary models are Gemma-2-2B-it~\citep{gemmateam2024gemma2}, Qwen3-1.7B~\citep{yang2025qwen3}, Llama3.2-3B~\citep{grattafiori2024llama3} and Qwen3-4B~\citep{yang2025qwen3}. 

\paragraph{Test data.} For \textbf{German} the per-prompt experiments use $30$ test English prompts from the Dolly dataset~\citep{databricks2023dolly}, which are disjoint from the contrast pairs used to build the steering vector (Section~\ref{sec:steering-vectors}). The universal experiments train
a single suffix on $20$ prompts and test the suffix on $30$ further prompts. For \textbf{refusal}, we utilise the harmful test set of \citet{arditi2024refusal} in two ways.
The \emph{breadth} split takes one prompt from each of the $22$ categories in the dataset, resulting in $22$ total prompts. The \emph{depth} split instead takes $5$ prompts from each of $4$ training categories, with a disjoint set of $5$ other prompts from the same $4$ categories as test, as well as an additional set of $5$ prompts from $4$ categories that are distinct from the training ones ($60$ prompts in total). 

\paragraph{Judging.}

We judge \textbf{German} success rates with the \texttt{langdetect} library~\citep{danilak2014langdetect}, additionally checking whether the response consists of degenerate repetitive phrases. We note that a short response that is genuinely German can still be misclassified when it contains English names or titles. A result is German if and only if the text is non-repetitive and the library gives ``de'' as output. To judge \textbf{refusal} success rates, following \citet{arditi2024refusal}, we first check the model responses across a set of phrases, i.e. whether a part of the output and a phrase match together (e.g. ``I cannot help with that''). We then also check for safety, where we use the HarmBench-Llama-2-13b classifier~\citep{mazeika2024harmbench} as the model judge. We acknowledge the difficulty in objective scoring for safety, and settled on this judge after comparing it against alternatives such as \texttt{gpt-5.5}, \texttt{claude-opus-4.8}, and \texttt{HarmBench-Mistral-7b-val-cls}. For Claude and GPT, we found issues in obtaining a score from the models due to safety classifiers, and we preferred the Llama classifier over Mistral as it is more widely used in the literature. We treat the HarmBench judge as our main refusal score, as otherwise the phrase based metric scores roughly twice as high as Harmbench (page~\pageref{par:calibration}).

\paragraph{Comparisons.}

We compare ActGCG against three alternatives. \emph{Soft suffixes} optimise directly over continuous embeddings that do not necessarily correspond to words, where we use gradient descent with Adam~\citep{kingma2015adam} to bring the activations as close as possible to the same activation matching loss of Equation~\ref{eq:ares-loss}. Since the optimiser explores a continuous space rather than the discrete vocabulary, a soft suffix can reach the target more easily than a discrete one. The original \emph{GCG} method~\citep{zou2023gcg} minimises the
cross entropy towards a fixed string (an affirmative string for refusal, a German string for the language task), determining what a discrete suffix can do \emph{without} any steering vector. \emph{SSR} steering (Equation~\ref{eq:ssr-loss}) is the closest prior method, matching the steering direction with a cosine loss rather than the full target activation.

\paragraph{Default settings.} Unless a subsection states otherwise, every suffix is $20$ tokens initialised to \texttt{x x \ldots x} and optimised for $250$ steps, with search width $512$, top-$k = 256$, and a single token replaced per step (as in Section~\ref{sec:methodology}). The target activations are built at the four layers of Table~\ref{tab:layers} with steering strength $\alpha = 2.0$ for German and $\alpha = 1.0$ for refusal. The per-prompt, universal and model translation ActGCG runs use a fixed seed of $0$ (except for the seed sweep in Section~\ref{sec:setup}). Each subsection below reports only its deviations from these defaults.

\chapter{Experiments}%
\label{sec:experiments}

\section{Different values of $\alpha$ for German}

 We test different values of $\alpha$ to change the strength of steering vectors added in the four chosen layers for a given model. We use ActGCG to optimise a suffix towards the target activations for each of the 30 German test prompts. We sweep $\alpha \in \{0.4, 0.7, 1.0, 1.3, 1.6, 2.0, 3.0, 4.0, 5.0\}$ on Gemma-2-2B and Qwen3-1.7B, and run Llama-3.2-3B and Qwen3-4B at $\alpha = 2.0$ only.

\begin{figure}[htbp]
  \centering
  \includegraphics[width=0.9\linewidth]{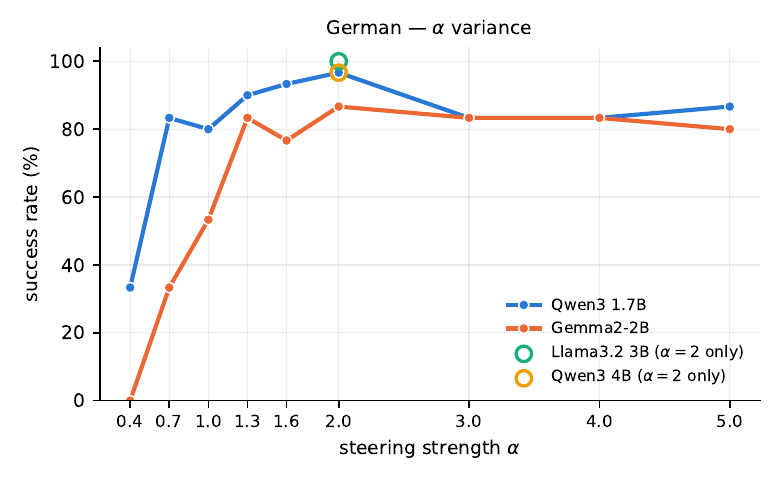}
  \caption{Per-prompt German success rate (langdetect) versus the steering strength $\alpha$, for
           each of the four main models on $30$ prompts. Each $\alpha$ scales the steering vector
           that defines the activation target.}
  \label{fig:alpha-models}
\end{figure}

We find that $\alpha = 2$ is a good value for the German steering vector across the four chosen layers, as it gives the highest success rate. The suffix optimised on the German steering vector reliably induces the same behaviour.

\section{Different seeds for different values of alpha}
\label{sec:setup}

We found some variance in results when running an experiment more than once since the original GCG method is stochastic. To minimise the effect of randomness in our results, we repeat our $\alpha$ sweep at $\alpha \in \{1.0, 1.3, 2.0, 3.0\}$ for five different seeds ($0$--$4$) on Gemma-2-2B. We also fix the default seed to $0$ in our ActGCG experiments. 

\begin{figure}[htbp]
  \centering
  \includegraphics[width=0.8\linewidth]{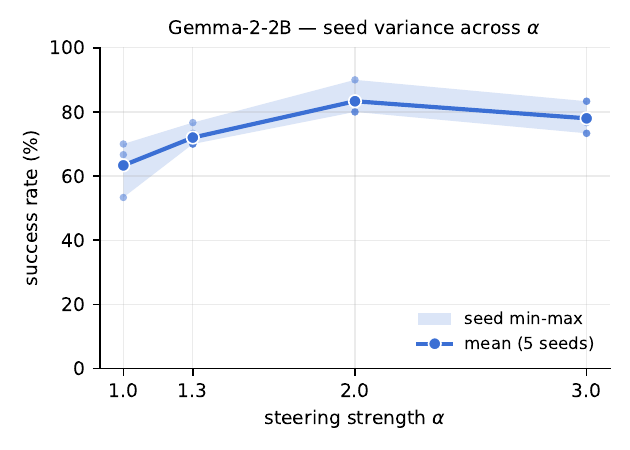}
  \caption{Per-prompt German success rate on $30$ prompts across $\alpha$, with five different seeds on Gemma-2-2B.}
  \label{fig:seeds}
\end{figure}

Despite the variance from GCG's randomness, $\alpha = 2$ remains the strongest choice overall.

\section{Suffix lengths}

We now vary the number of optimised tokens in the suffix, which we otherwise fix at $20$. On Gemma-2-2B  we sweep the suffix length over
$\{5, 10, 15, 20, 25, 50\}$ tokens, repeating each length for five seeds ($0$--$4$). We test on $n = 30$ German prompts.

\begin{figure}[htbp]
  \centering
  \includegraphics[width=0.8\linewidth]{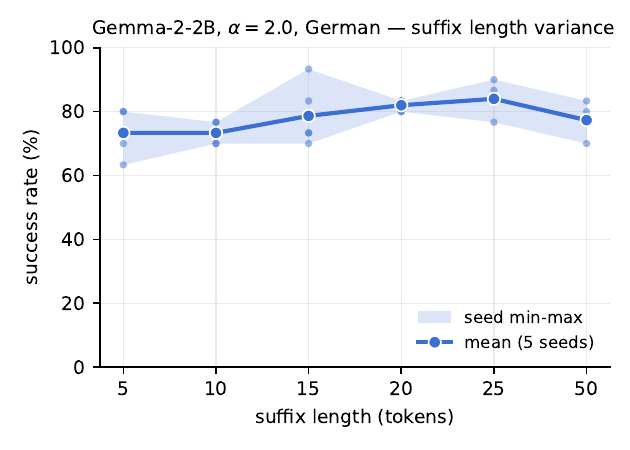}
  \caption{Per-prompt German success rate on 30 prompts for different suffix lengths, with one point per seed and the mean.}
  \label{fig:suffix-len}
\end{figure}

We find that changing the suffix length can increase the success rate. At the same time, variance is low at suffix token length 20, while increasing suffix length costs more optimisation time but minimal gain. Hence, we kept our experiments to have the same suffix length of 20.

\section{Checking target on German}
\label{sec:target-german}

We check whether the activation target itself yields German answers. We do this by adding the German steering vector ($\alpha = 2.0$) directly to the residual stream at the filler positions of the four target layers (Table~\ref{tab:layers}) \emph{before generation} (with no suffix involved) using Gemma-2-2B, and measuring the German rate on the same $30$ prompts.

\begin{table}[htbp]
  \centering
  \caption{Steering vector added at the filler positions, the target which ActGCG optimises towards (Gemma-2-2B, $\alpha = 2.0$, $30$ prompts).}
  \label{tab:sv-baseline}
  \begin{tabular}{l r}
    \toprule
    Method & {Success \pct} \\
    \midrule
    Original SV on filler tokens & 90.00 \\
    \bottomrule
  \end{tabular}
\end{table}

Our target is worth striving towards; however, the result itself is not 100\%. Therefore, the success of our method is capped by the success of the target.

\section{Comparing results to the original GCG method and SSR for German}
\label{sec:compare-german}

We compare our method to the original GCG algorithm using the same per-prompt setup. Instead of matching the target activations with a steering vector, the original GCG minimises the cross-entropy loss towards a fixed target string. We determine the target string to be ``Hier ist meine Antwort auf Deutsch:'' (``Here is my answer in German:''). Otherwise, we use the same $n = 30$ German Dolly prompts, same initialisation, and the same $250$ steps.

\begin{table}[htbp]
  \centering
  \caption{Comparing ActGCG to the original GCG method with the same per-prompt setup apart from the loss.}
  \label{tab:compare-gcg-german}
  \begin{tabular}{l r r}
    \toprule
    Model & Original GCG & ActGCG \\
    \midrule
    Gemma-2-2B  & 36.7 & 86.7 \\
    Qwen3-1.7B  & 83.3 & 96.7 \\
    \bottomrule
  \end{tabular}
\end{table}

We notice that ActGCG performs better for the German concept on both models. We note that this experiment could be further improved by sweeping a set of German phrases for the target string.

Besides GCG, we also compare our method to SSR~\citep{winninger2025ssr} due to the similarity in goals. Instead of minimising the squared distance to the target activations, SSR drives the cosine similarity between the steering vector and the activations towards a target value (Equation~\ref{eq:ssr-loss}). The original implementation reads the last prompt token, which differs from our method. Therefore, we run SSR twice; first by faithfully following its single token position, and second by reading the filler positions the same way our method does. We run it on Gemma-2-2B, and the rest of the setup is the same.

\begin{table}[htbp]
  \centering
  \caption{Comparing ActGCG to the SSR method with the same per-prompt setup and $\alpha = 2.0$}
  \label{tab:ssr-german}
  \begin{tabular}{l r}
    \toprule
    Method & German \pct \\
    \midrule
    SSR cosine, faithful & 40.0 \\
    SSR cosine, filler          & 53.3 \\
    ActGCG              & 86.7 \\
    \bottomrule
  \end{tabular}
\end{table}

For Gemma-2-2B, our method outperforms both variants. This indicates that the cosine objective is weaker than our $L_2$ loss. We also see that reading at the filler positions performs better than the single token.

\section{Universal suffixes on German}
\label{sec:univ-german}

We now optimise a single \emph{universal} suffix that steers many prompts at once, rather than having one suffix per prompt. Following Section~\ref{sec:universal-mt}, a suffix is trained jointly on $20$ German prompts, and then evaluated on the same $20$ training prompts as well as a set of $30$ test prompts. We also transfer the finished suffix from the source model to a different model, evaluated on the $20$ training prompts. Where available, we transfer to a similarly sized (or smaller) and a slightly larger model of the same family.

\begin{table}[htbp]
    \centering
    \caption{Universal German suffix, with $\alpha = 2.0$ and $20$ train prompts. Training and
             test are for the source model; transfer applies the same suffix to another model,
             evaluated on the training prompts.}
    \label{tab:universal-german}
    \begin{tabular}{l l r}
      \toprule
      Source model & Evaluation & German \pct \\
      \midrule
      \multirow{4}{*}{Gemma-2-2B}
        & training (20)   & 90.0 \\
        & test (30)       & 73.3 \\
        & Gemma-1.1-2B    & 65.0 \\
        & Gemma-2-9B      &  5.0 \\
      \midrule
      \multirow{4}{*}{Qwen3-1.7B}
        & training (20)   & 95.0 \\
        & test (30)       & 80.0 \\
        & Qwen2.5-1.5B    & 95.0 \\
        & Qwen3-4B        & 60.0 \\
      \midrule
      \multirow{3}{*}{Llama-3.2-3B}
        & training (20)   & 95.0 \\
        & test (30)       & 96.7 \\
        & Llama-3.2-1B    & 95.0 \\
      \bottomrule
    \end{tabular}
  \end{table}

We find that the test set generally performs similarly to or worse than the training set. For Llama-3.2-3B, both sets only had one failed prompt, but the test set showed a slight increase, likely due to larger number of prompts tested. Transferring the suffix to a similarly sized or smaller model performed better than transferring it to a larger model.

\section{Model translation for universal suffixes on German}
\label{sec:mt-german}

The universal suffix is mostly not interpretable, so we check if a model translation preserves the meaning of the concept better.  We run the universal steps as usual for the first half, and then insert a \emph{model-translation} step in the middle of the steps (at step $125$) where we make one call to a \emph{proposer} model that rewrites the suffix into fluent text. We try two proposers, GPT-5.5 and Llama-3.1-8B. For the remaining steps, we continue as usual. We use the same $20$ training and $30$ test prompts. We also check a variant \texttt{end\_ask} where we ask at the end instead of the middle. Suffixes transferred to other models are evaluated on the $30$ test prompts (in comparison, cross-model transfers in Section~\ref{sec:univ-german} were checked on the easier training prompts).

\begin{table}[htbp]
  \centering
  \caption{Universal modified GCG with model translation, German ($\alpha = 2.0$,
           $20$ train, $30$ test prompts). The proposer rewrites the suffix in the middle of the steps (step $125$); \texttt{end\_ask} rewrites only after the final step instead. Transferred suffixes are evaluated on the $30$ test prompts.}
  \label{tab:gcgloop-translate-german}
  \footnotesize
  \setlength{\tabcolsep}{4pt}
  \begin{tabular}{l l r r r r r}
    \toprule
    & & \multicolumn{2}{c}{Source model} & \multicolumn{2}{c}{Transfer} & \\
    \cmidrule(lr){3-4}\cmidrule(lr){5-6}
    Source model & Proposer & Training & Test & Gemma-1-2B & Gemma-2-9B
      & end\_ask \\
    \midrule
    \multirow{2}{*}{Gemma-2-2B}
      & GPT-5.5      & 95.0 & 96.7  & 83.3 & 100.0 & 70.0 \\
      & Llama-3.1-8B & 85.0 & 83.3  & 83.3 & 100.0 &  0.0 \\
    \midrule
    Llama-3.2-3B
      & GPT-5.5      & 100.0  & 100.0 & \na  & \na   & 93.3 \\
    \bottomrule
  \end{tabular}
\end{table}

We find that translating the suffix with a model in the middle of the steps performs better in inducing German behaviour than using no translation step, as well as in generalising to new prompts and transferring to other models. 

\section{Tracking the change of loss for German}
\label{sec:track-german}

We log the cosine similarity (between the suffix's filler token activations and the target activations) and the $L_2$ distance to the target every $25$ steps at each of the four target layers, for both the universal suffix and the model-translation run.
  
\begin{figure}[htbp]
  \centering
  \includegraphics[width=0.9\linewidth]{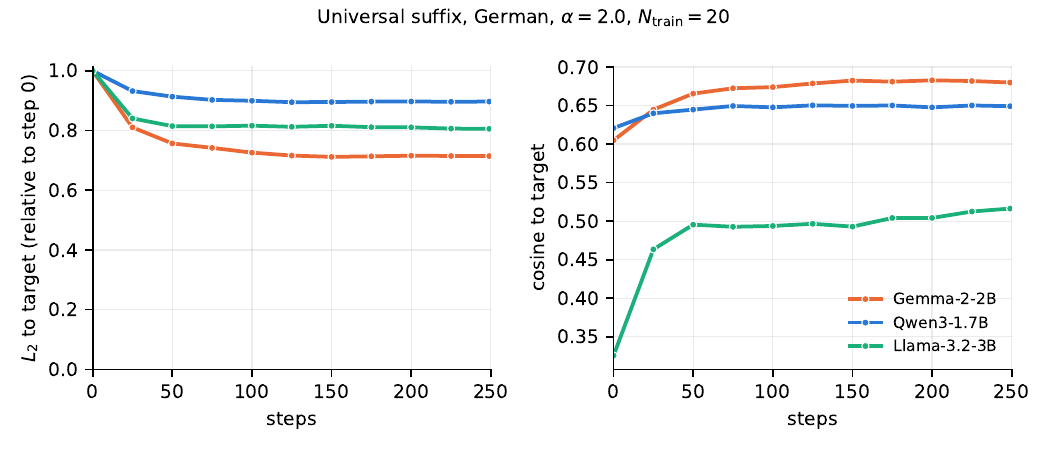}
  \caption{$L_2$ distance to the target activations (normalised to its step-0 value) and
           cosine similarity between the filler-token activations and the target, every $25$ steps and averaged across the four target
           layers, for the universal German run with $\alpha = 2.0$ and $20$ training prompts.}
  \label{fig:uni_loss}
\end{figure}

\begin{figure}[htbp]
  \centering
  \begin{subfigure}{0.32\linewidth}\includegraphics[width=\linewidth]{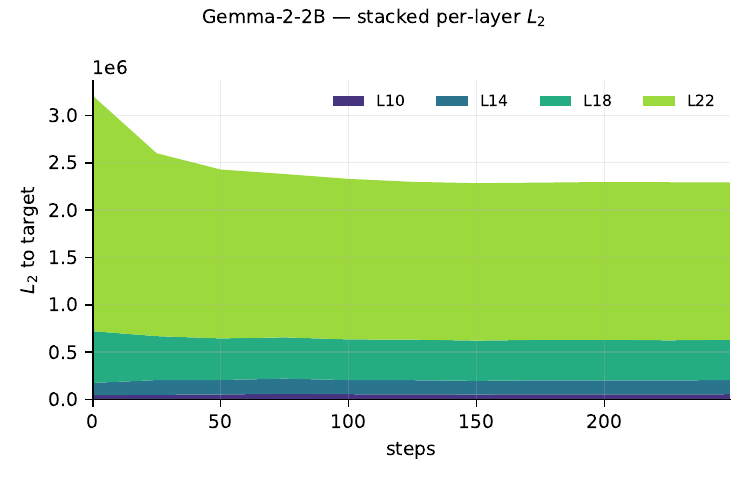}\end{subfigure}\hfill
  \begin{subfigure}{0.32\linewidth}\includegraphics[width=\linewidth]{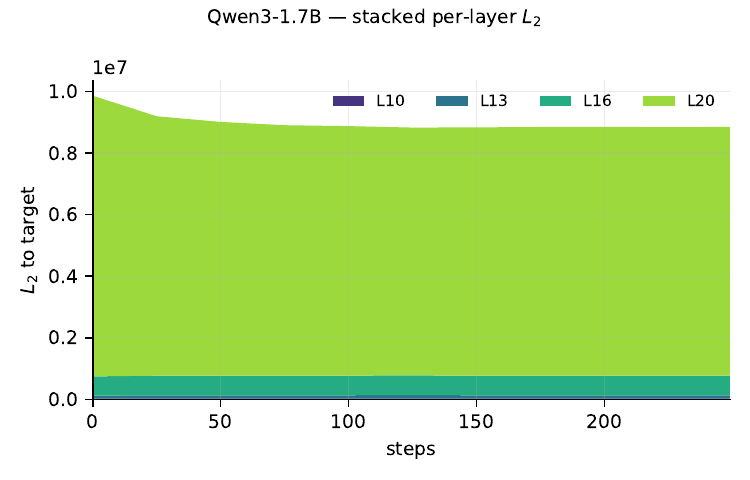}\end{subfigure}\hfill
  \begin{subfigure}{0.32\linewidth}\includegraphics[width=\linewidth]{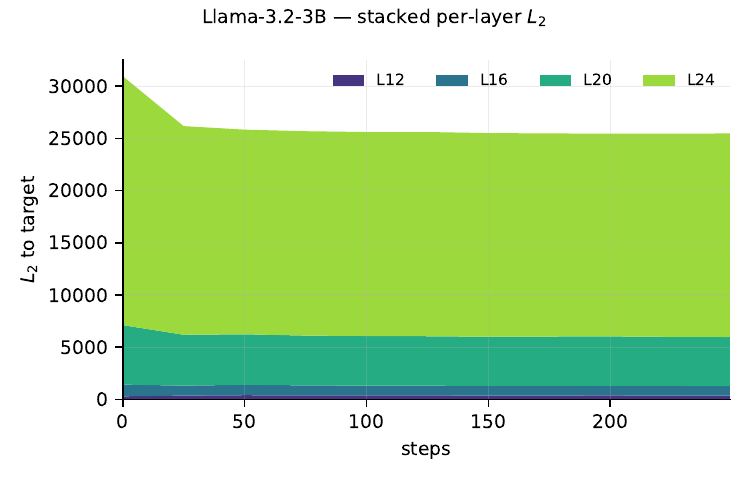}\end{subfigure}
  \caption{Per-layer activation matching for the universal German suffix ($\alpha = 2.0$). The four target layers' $L_2$ distances to the target, stacked.}
  \label{fig:uni_stack_german}
\end{figure}

\begin{figure}[htbp]
  \centering
  \begin{subfigure}{0.32\linewidth}\includegraphics[width=\linewidth]{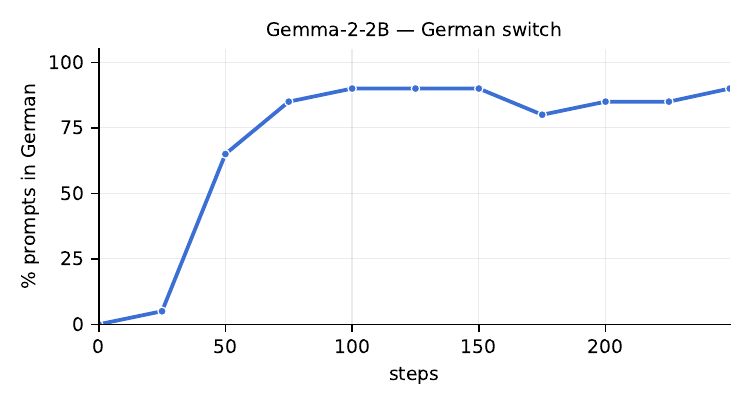}\end{subfigure}\hfill
  \begin{subfigure}{0.32\linewidth}\includegraphics[width=\linewidth]{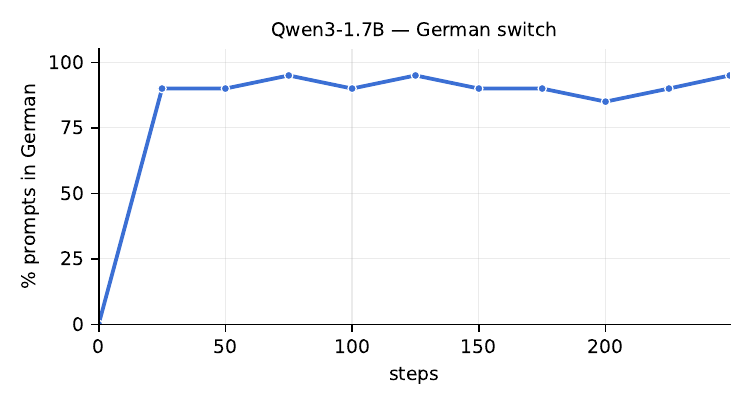}\end{subfigure}\hfill
  \begin{subfigure}{0.32\linewidth}\includegraphics[width=\linewidth]{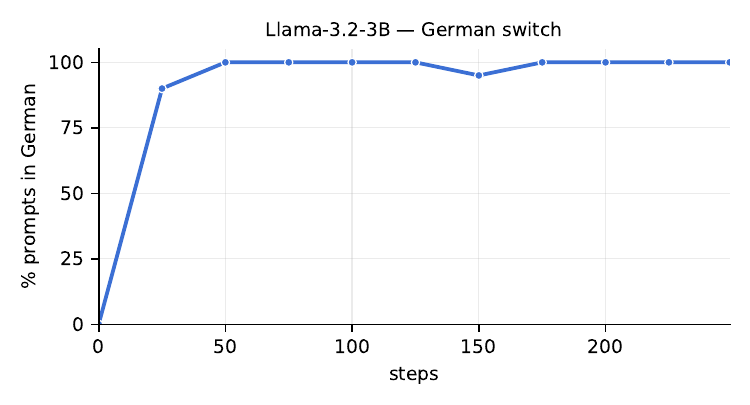}\end{subfigure}
  \caption{Fraction of train prompts answered in German every $25$ steps, for the universal German suffix.}
  \label{fig:uni_switch_german}
\end{figure}

\paragraph{Universal suffix.} We check how the cosine similarity and the $L_2$ distance loss changes throughout the ActGCG steps. We log both every $25$ steps at each of the four target layers, for universal suffix experiments (Figure~\ref{fig:uni_loss}). Figure~\ref{fig:uni_loss} sums both losses over the layers and Appendix~\ref{app:perlayer-cosine} considers the cosine loss for each of the chosen layers (with one target per layer). Figure~\ref{fig:uni_stack_german} shows the contribution of each chosen layer to the $L_2$ loss. Figure~\ref{fig:uni_switch_german} considers the change in German rate for the given suffix every $25$ steps. We notice overall that the loss decreases and the results get closer to the target. However, we also see that we get a roughly logarithmic curve, and increasing the number of steps for longer than 100 steps doesn't significantly change the loss. The same descent holds for the individual per-prompt runs (Appendix~\ref{app:loss-grid}).

\begin{figure}[htbp]
  \centering
  \includegraphics[width=0.9\linewidth]{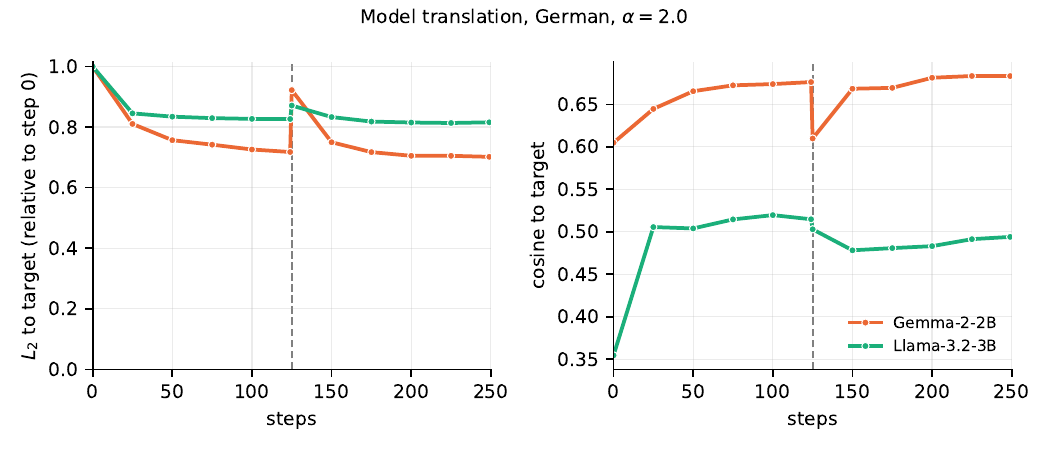}
  \caption{Model-translation German run: $L_2$ (normalised to step 0) and cosine to the target
           over steps, averaged across layers. The dashed line marks the mid-run proposer
           rewrite.}
  \label{fig:translate_loss_german}
\end{figure}

\begin{figure}[htbp]
  \centering
  \begin{subfigure}{0.48\linewidth}\includegraphics[width=\linewidth]{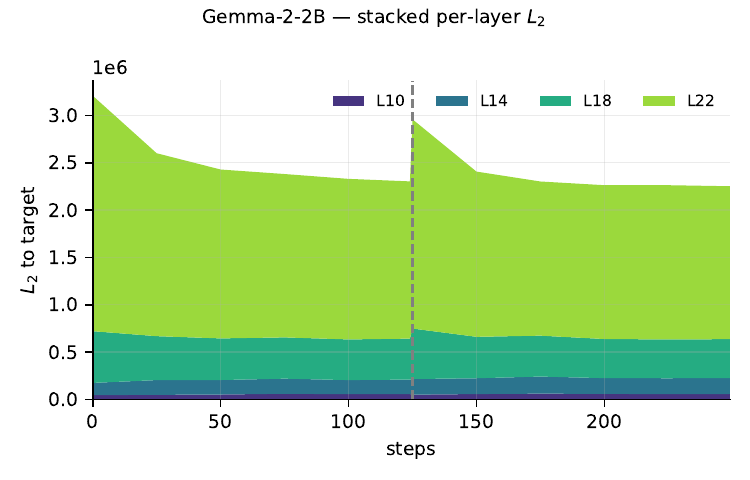}\end{subfigure}\hfill
  \begin{subfigure}{0.48\linewidth}\includegraphics[width=\linewidth]{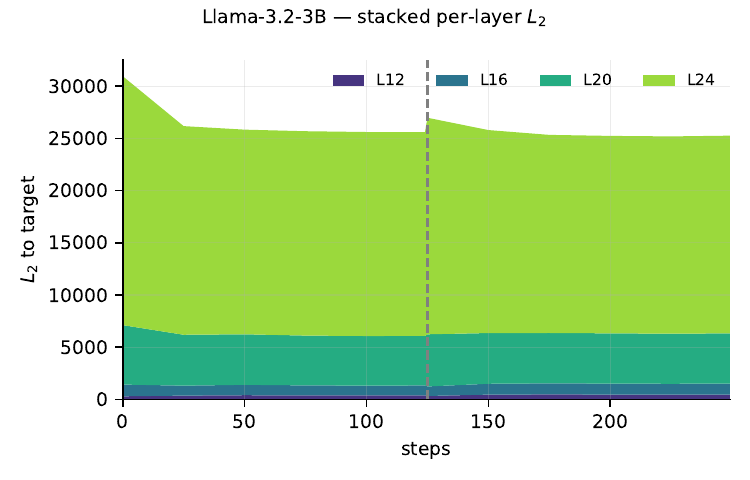}\end{subfigure}
  \caption{Per-layer activation matching for the \emph{model-translation} German run for $\alpha = 2.0$. Stacked per-layer $L_2$ distances; the dashed line is the step-125 proposer rewrite.}
  \label{fig:mt_stack_german}
\end{figure}

\begin{figure}[htbp]
  \centering
  \begin{subfigure}{0.48\linewidth}\includegraphics[width=\linewidth]{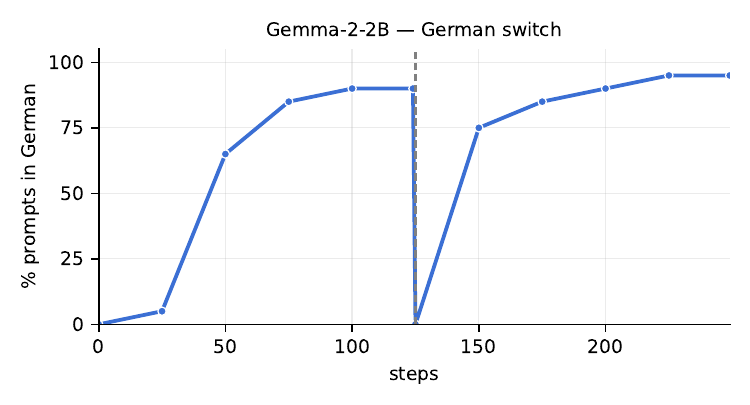}\end{subfigure}\hfill
  \begin{subfigure}{0.48\linewidth}\includegraphics[width=\linewidth]{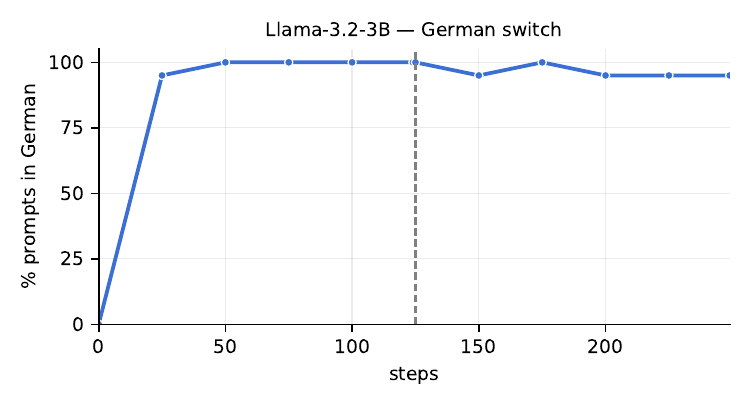}\end{subfigure}
  \caption{Fraction of train prompts answered in German every $25$ steps for the model-translation
           run.}
  \label{fig:mt_switch_german}
\end{figure}

\FloatBarrier
\paragraph{Model translation.} We repeat the same procedure for the model-translation run. Figure~\ref{fig:translate_loss_german} sums the $L_2$ distance and averages the cosine over the layers, and Figure~\ref{fig:mt_stack_german} breaks the $L_2$ down per layer. Figure~\ref{fig:mt_switch_german} tracks the German rate over the run. We observe a sharp change in loss when the suffix is translated by the proposer model. Despite the fact that model translation experiments produce higher success results, we do not observe a significant change in loss at the end of the steps. Nevertheless, we believe the re-initialisation in the middle of the steps might drive the model to explore new suffixes better.

\section{Refusal}

We now apply ActGCG to the anti-refusal direction rather than German, on $22$ harmful prompts (using the breadth split, one prompt from each category). For every prompt, we optimise a single suffix. We sweep for $\alpha \in \{0.5, 1.0, 1.5, 2.0\}$, and compare against a ``None'' baseline where we don't optimise for any suffix. The jailbreak rates are judged with HarmBench-Llama-2-13b. We note that $\alpha$ values are negated in the code for the anti direction.

\begin{figure}[htbp]
    \centering
    \includegraphics[width=0.85\linewidth]{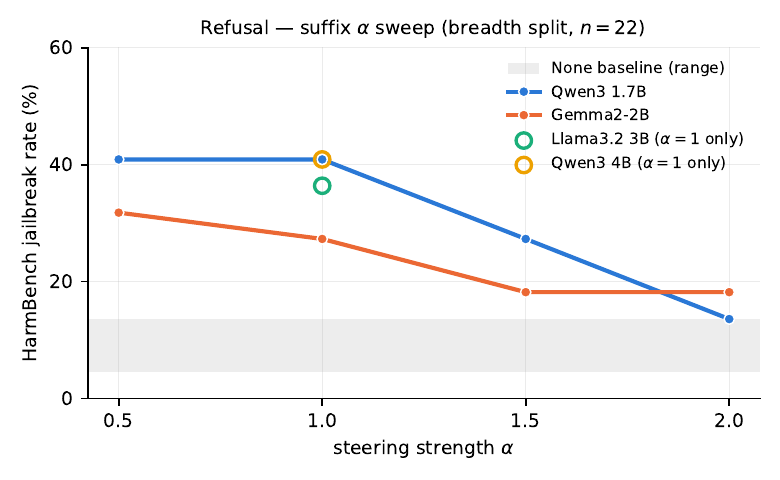}
    \caption{Anti-refusal direction, breadth split, $22$ prompts, judged with HarmBench. The grey band marks the unsteered baseline (\texttt{None}) range across the four
             models ($4.5$--$13.6$\%).}
    \label{fig:refusal-ares}
  \end{figure}

The suffix induces jailbreaking more often than the unsteered baseline, which indicates that our optimisation induces the anti-refusal behaviour. At the same time, the steering strength is much weaker than for German (Figure~\ref{fig:alpha-models}). 

On the suffix itself, Gemma-2-2B and Qwen3-1.7B together do slightly better at $\alpha = 0.5$ than at $\alpha = 1.0$, and the rate then falls as $\alpha$ grows further. We nonetheless adopt $\alpha = 1.0$ as the default, since the reachable target itself peaks at $\alpha \in [1.0, 1.5]$ while $\alpha = 0.5$ is too weak (Section~\ref{sec:target-refusal}). The suffix still matches the target well at $\alpha = 1.0$ but degrades beyond it.

\section{Checking our target for refusal}
\label{sec:target-refusal} 

We check whether the activation target itself yields jailbreaking answers, as we did for the German direction. We add the refusal steering vector for $\alpha \in \{0.5, 1.0, 1.5, 2.0\}$ directly into the residual stream at the filler positions of the four target layers (Table~\ref{tab:layers}) \emph{before generation} on Gemma-2-2B, with no suffix involved, and measure both metrics on the same $22$ breadth-split prompts (one per category).

\begin{table}[htbp]
    \centering
    \caption{Refusal steering vector applied at the filler positions, the target
             ActGCG optimises towards (Gemma-2-2B, breadth split, $n = 22$). HarmBench jailbreak
             rate and phrase-based anti-refusal rate \pct.}
    \label{tab:refusal-target}
    \begin{tabular}{r r r}
      \toprule
      {$\alpha$} & {HarmBench \pct} & {Phrase-based \pct} \\
      \midrule
      0.5 & 22.73 & 63.64 \\
      1.0 & 50.00 & 81.82 \\
      1.5 & 50.00 & 90.91 \\
      2.0 & 40.91 & 77.27 \\
      \bottomrule 
    \end{tabular} 
\end{table}

The HarmBench target peaks at $\alpha \in [1.0, 1.5]$, while the phrase-based rate is roughly twice as high\label{par:calibration}. The target never reaches $100\%$, hence the success of our ActGCG method is capped by it. It is also far weaker than its German counterpart ($90.00\%$, Section~\ref{sec:target-german}), which may be the reason for the lower refusal success rates throughout our experiments.

\section{Comparing results to the original GCG method and SSR for refusal}

Similar to Section~\ref{sec:compare-german}, we now compare our method to the original GCG algorithm with the same per-prompt setup for refusal. We use the same $n = 22$ prompts with the same suffix initialisation, as well as the same number of $250$ steps. We compare for both the phrase based metric and the safety metric for jailbreaking.

\begin{table}[htbp]
    \centering
    \caption{Per-prompt refusal, breadth split ($22$ prompts), comparing the original GCG against ActGCG  with $\alpha = 1.0$. HarmBench jailbreak rate and the phrase-based
             anti-refusal rate, both in \%.}
    \label{tab:compare-gcg-refusal}
    \begin{tabular}{l r r r r}
      \toprule
      & \multicolumn{2}{c}{Original GCG} & \multicolumn{2}{c}{ActGCG} \\
      \cmidrule(lr){2-3}\cmidrule(lr){4-5}
      Model & HarmBench & Phrase & HarmBench & Phrase \\
      \midrule
      Gemma-2-2B  & 13.6 & 36.4  & 27.3 & 90.9 \\
      Qwen3-1.7B  & 50.0 & 100.0 & 40.9 & 90.9 \\
      \bottomrule
    \end{tabular}
  \end{table}

Considering both the safety and the phrase metric, our ActGCG performs stronger than the original GCG for Gemma-2-2B, but worse for Qwen3-1.7B. Overall, our activation target matches the fixed string target of GCG, but does not show a major improvement.

We compare our method to SSR~\citep{winninger2025ssr} again, using the same last token and filler token variants as before. Apart from the change in loss, we use the same per-prompt setup as before.

\begin{table}[htbp]
    \centering
    \caption{Per-prompt anti-refusal results on Gemma-2-2B, breadth split with $22$ prompts. Comparing both SSR variants against ActGCG. HarmBench jailbreak rate and the phrase-based
             anti-refusal rate, both in \%.} 
    \label{tab:ssr-refusal}
    \begin{tabular}{l r r} 
      \toprule
      Method & HarmBench & Phrase \\ 
      \midrule
      SSR cosine, faithful & 4.5  & 100.0 \\
      SSR cosine, filler   & 22.7 & 100.0 \\
      ActGCG               & 27.3 & 90.9 \\
      \bottomrule
    \end{tabular} 
  \end{table}

We observe that ActGCG performs better than both SSR variants on the safety metric. At the same time, for the phrase based metric, ActGCG nevertheless performs slightly worse. 

\section{Universal suffixes on refusal}
\label{sec:univ-refusal}

We now optimise a single \emph{universal} suffix that steers many prompts at once, rather than
having one suffix per prompt. We evaluate in two ways. First, we do a ``breadth split'', where we
train on $22$ prompts each from a different category, and then test on $22$ new prompts from all
different categories again. Then, we do a ``depth split'' (five prompts from each of four training
categories, then five prompts from the same categories for test, and finally five prompts from four new categories). We also transfer to a similarly sized (or smaller) and a slightly larger model of the same family, evaluated on training prompts for both split variants.

\begin{table}[htbp]
  \centering                     
  \caption{Universal suffix on refusal, breadth split ($22$ prompts, one prompt per category).
           HarmBench jailbreak rate and phrase-based anti-refusal rate, both in \%.}
  \label{tab:universal-refusal}
  \begin{tabular}{l l r r}
    \toprule
    Source model & Evaluation & HarmBench & Phrase \\
    \midrule
    \multirow{4}{*}{Gemma-2-2B}
      & training  & 45.5 & 95.5 \\
      & test     & 31.8 & 63.6 \\
      & Gemma-1.1-2B  & 9.1  & 22.7 \\
      & Gemma-2-9B    & 0.0  & 4.5  \\
    \midrule
    \multirow{4}{*}{Qwen3-1.7B}
      & training  & 45.5 & 100  \\
      & test     & 59.1 & 100  \\
      & Qwen2.5-1.5B  & 31.8 & 50.0 \\
      & Qwen3-4B      & 9.1  & 22.7 \\
    \midrule
    \multirow{3}{*}{Llama-3.2-3B}
      & training  & 36.4 & 86.4 \\
      & test     & 36.4 & 81.8 \\
      & Llama-3.2-1B  & 4.5  & 18.2 \\
    \bottomrule
  \end{tabular}
\end{table}

\begin{table}[htbp]
  \centering                       
  \caption{Universal suffix on refusal, depth split. HarmBench jailbreak rate and phrase-based
           anti-refusal rate, both in \%.}
  \label{tab:universal-refusal-depth}
  \begin{tabular}{l l r r}
    \toprule
    Source model & Evaluation & HarmBench & Phrase \\
    \midrule
    \multirow{5}{*}{Gemma-2-2B}
      & training     & 45.0 & 90.0 \\
      & test         & 30.0 & 85.0 \\
      & new categories      & 30.0 & 60.0 \\
      & Gemma-1.1-2B & 5.0  & 10.0 \\
      & Gemma-2-9B   & 10.0 & 10.0 \\
    \midrule
    \multirow{5}{*}{Qwen3-1.7B}
      & training     & 35.0 & 100  \\
      & test         & 45.0 & 95.0 \\
      & new categories      & 35.0 & 100  \\
      & Qwen2.5-1.5B & 40.0 & 100  \\
      & Qwen3-4B     & 25.0 & 50.0 \\
    \midrule
    \multirow{4}{*}{Llama-3.2-3B}
      & training     & 65.0 & 85.0 \\
      & test         & 60.0 & 90.0 \\
      & new categories      & 50.0 & 80.0 \\
      & Llama-3.2-1B & 10.0 & 10.0 \\
    \bottomrule
  \end{tabular}
\end{table}

\FloatBarrier
We notice that the phrase based metric scores significantly higher than the safety metric. The
training prompts generally score higher than the test prompts (except for slight increase in
Qwen3-1.7B), and transfers to smaller or similarly sized models perform better than larger models,
similar to the German transfers (Section~\ref{sec:univ-german}). We also observe that the suffix does generalise to unseen categories.

\section{Model translation for universal suffixes on refusal}

We now evaluate a variant of the universal suffix again, where we insert a model-translation in the
middle of the steps. We namely call a proposer model that rewrites the suffix into coherent text as
in Section~\ref{sec:mt-german}. As proposers, we utilise GPT-5.5 and Llama-3.1-8B again. We compare
for both the depth and breadth split variants. The transfer models are evaluated on the test prompts (same as in Section~\ref{sec:mt-german}).

\begin{table}[htbp]
  \centering
  \caption{Breadth split ($n = 22$, one prompt per category). Model-translation universal suffix on
           refusal with HarmBench jailbreak rate and phrase-based anti-refusal rate, both in \%.}
  \label{tab:gcgloop-refusal}
  \begin{tabular}{l l l r r}
    \toprule
    Source model & Proposer & Evaluation & HarmBench & Phrase \\
    \midrule
    \multirow{8}{*}{Gemma-2-2B}
      & \multirow{4}{*}{GPT-5.5}
          & training     & 45.5 & 86.4 \\
      &   & test        & 31.8 & 68.2 \\
      &   & Gemma-1.1-2B & 4.5  & 4.5  \\
      &   & Gemma-2-9B   & 4.5  & 13.6 \\
      \cmidrule(l){2-5}
      & \multirow{4}{*}{Llama-3.1-8B}
          & training     & 40.9 & 90.9 \\
      &   & test         & 22.7 & 72.7 \\
      &   & Gemma-1.1-2B & 0.0  & 9.1  \\
      &   & Gemma-2-9B   & 0.0  & 0.0  \\
    \midrule
    \multirow{3}{*}{Llama-3.2-3B}
      & \multirow{3}{*}{GPT-5.5}
          & training     & 36.4 & 81.8 \\
      &   & test         & 31.8 & 72.7 \\
      &   & Llama-3.2-1B & 4.5  & 4.5  \\
    \bottomrule
  \end{tabular}
\end{table}

\begin{table}[htbp]
  \centering
  \caption{Depth split ($20$ prompts each for training, test, new categories), model-translation universal suffix on refusal. HarmBench jailbreak rate and phrase-based anti-refusal rate, both in \%.}
  \label{tab:gcgloop-refusal-depth}
  \begin{tabular}{l l l r r}
    \toprule
    Source model & Proposer & Evaluation & HarmBench & Phrase \\
    \midrule
    \multirow{10}{*}{Gemma-2-2B}
      & \multirow{5}{*}{GPT-5.5}
          & training     & 10.0 & 85.0 \\
      &   & test          & 15.0 & 90.0 \\
      &   & new categories        & 5.0  & 60.0 \\
      &   & Gemma-1.1-2B & 10.0 & 15.0 \\
      &   & Gemma-2-9B   & 5.0  & 5.0  \\
      \cmidrule(l){2-5}
      & \multirow{5}{*}{Llama-3.1-8B}
          & training       & 55.0 & 95.0 \\
      &   & test          & 20.0 & 85.0 \\
      &   & new categories       & 15.0 & 85.0 \\
      &   & Gemma-1.1-2B & 10.0 & 10.0 \\
      &   & Gemma-2-9B   & 10.0 & 35.0 \\
    \midrule
    \multirow{4}{*}{Llama-3.2-3B}
      & \multirow{4}{*}{GPT-5.5}
          & training       & 45.0 & 80.0 \\
      &   & test           & 45.0 & 90.0 \\
      &   & new categories       & 40.0 & 80.0 \\
      &   & Llama-3.2-1B & 0.0  & 0.0  \\
    \bottomrule
  \end{tabular}
\end{table}

We find that in the case of refusal, a model translation in the middle of the steps generally
performs worse than the universal ActGCG with no translation. This is in contrast to our results for German, and shows that the German language may be easier to translate to a coherent meaning than the
more complex refusal concept.

\section{Tracking the change of loss for refusal}

We repeat the loss-tracking analysis of Section~\ref{sec:track-german} for refusal, on the breadth
split.

\begin{figure}[htbp]
  \centering
  \includegraphics[width=0.9\linewidth]{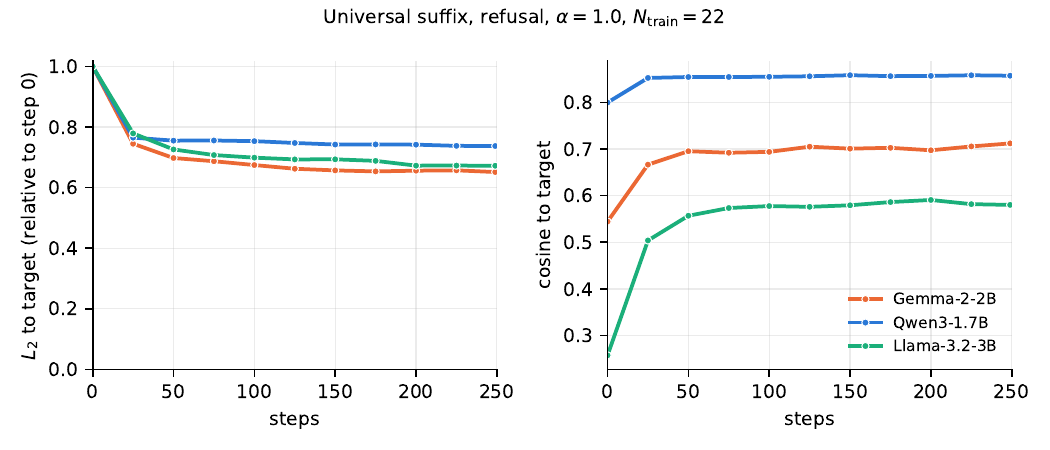}
  \caption{$L_2$ distance to the target activations (normalised to its step-0 value) and cosine
           similarity between the filler-token activations and the target, over optimisation steps
           and averaged across the four target layers, for the universal refusal run
           ($\alpha = 1.0$, $N_{\mathrm{train}} = 22$).}
  \label{fig:uni_loss_refusal}
\end{figure}

\paragraph{Universal suffix.} We first check the change of loss in our universal experiments every
$25$ steps. Figure~\ref{fig:uni_loss_refusal} sums the $L_2$ distance and averages the cosine over
the layers, Figure~\ref{fig:uni_stack_refusal} shows the per-layer contribution to the $L_2$ loss
(per-layer cosine in Appendix~\ref{app:perlayer-cosine}), and Figure~\ref{fig:uni_switch_refusal}
tracks the change in refusal rate for the phrase metric. The per-prompt runs show the same descent (Appendix~\ref{app:loss-grid}).

\begin{figure}[htbp]
  \centering
  \begin{subfigure}{0.32\linewidth}\includegraphics[width=\linewidth]{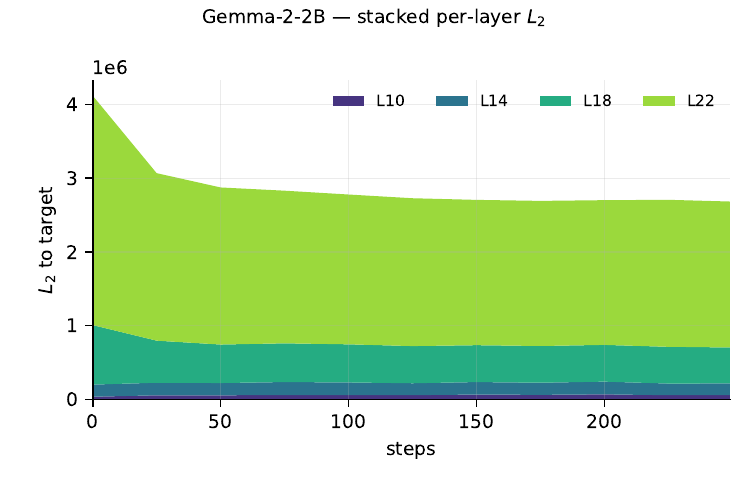}\end{subfigure}\hfill
  \begin{subfigure}{0.32\linewidth}\includegraphics[width=\linewidth]{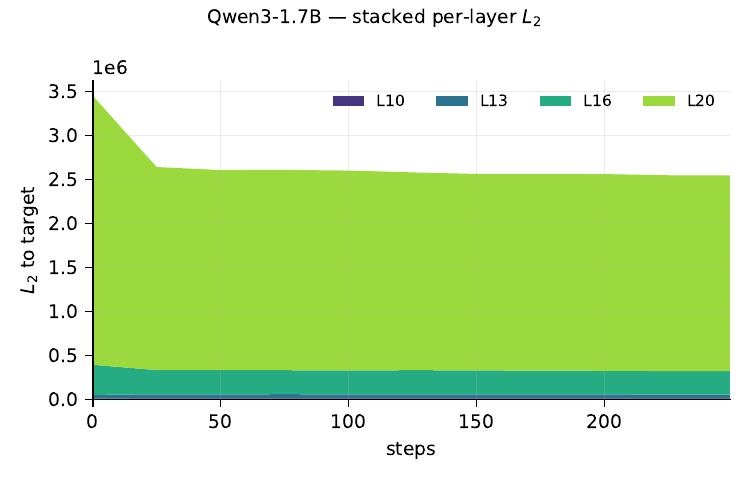}\end{subfigure}\hfill
  \begin{subfigure}{0.32\linewidth}\includegraphics[width=\linewidth]{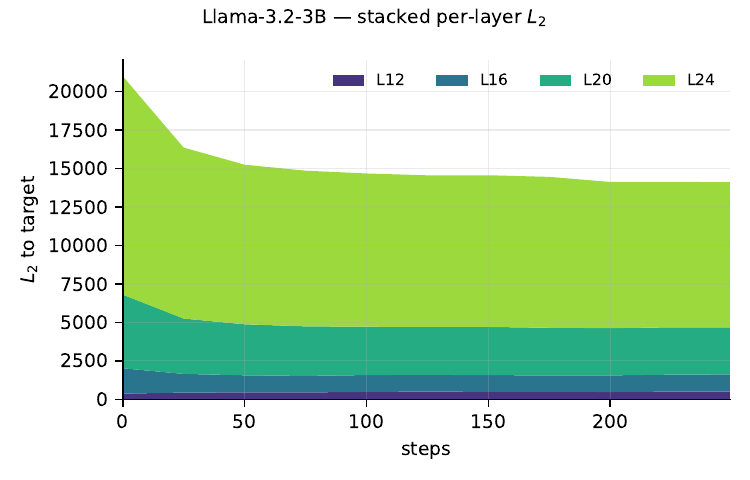}\end{subfigure}
  \caption{Stacked per-layer $L_2$ distance to the target for the universal anti-refusal suffix
           ($N_{\mathrm{train}} = 22$).}
  \label{fig:uni_stack_refusal}
\end{figure}

\begin{figure}[htbp]
  \centering
  \begin{subfigure}{0.32\linewidth}\includegraphics[width=\linewidth]{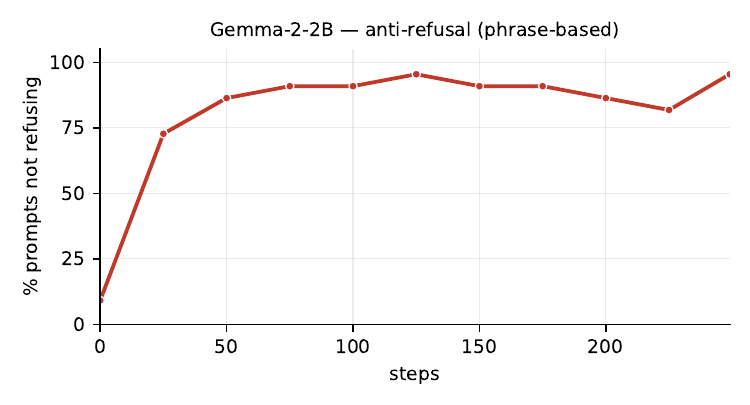}\end{subfigure}\hfill
  \begin{subfigure}{0.32\linewidth}\includegraphics[width=\linewidth]{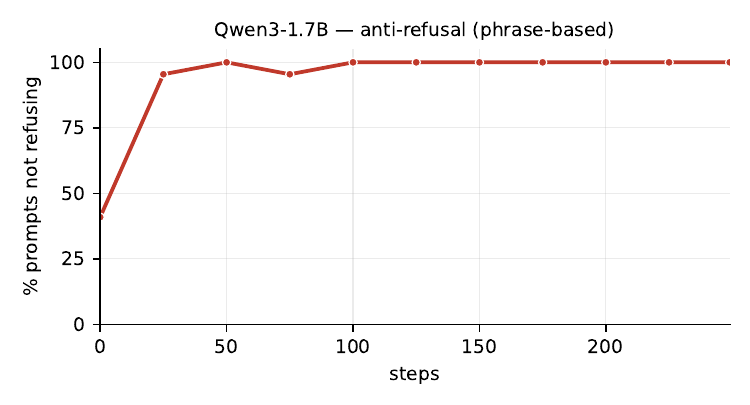}\end{subfigure}\hfill
  \begin{subfigure}{0.32\linewidth}\includegraphics[width=\linewidth]{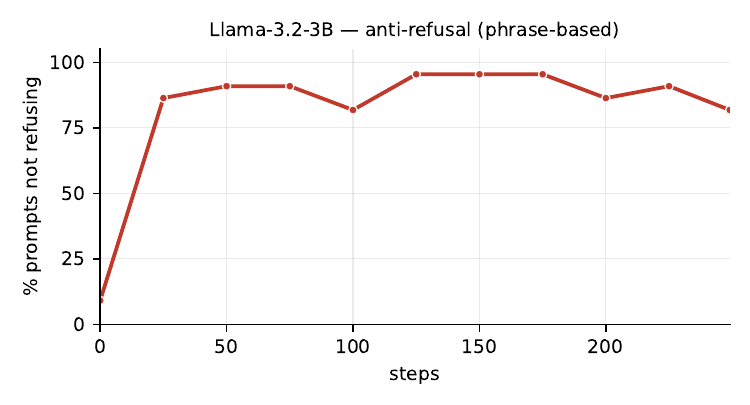}\end{subfigure}
  \caption{Fraction of train prompts whose output is \emph{not} a phrase-based refusal every $25$ steps, for the universal anti-refusal suffix.}
  \label{fig:uni_switch_refusal}
\end{figure}

\begin{figure}[htbp]
  \centering
  \includegraphics[width=0.9\linewidth]{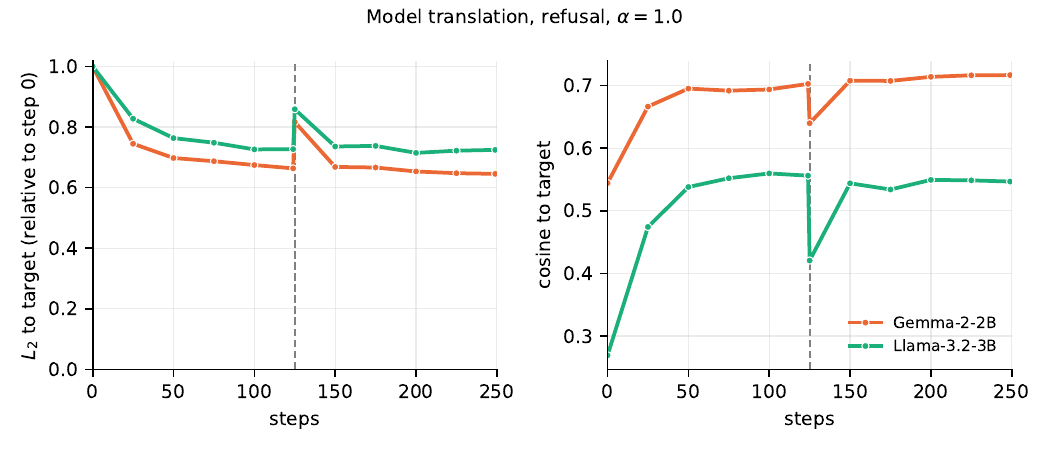}
  \caption{Model-translation refusal run. $L_2$ (normalised to step 0) and cosine to the target every $25$ steps, averaged across layers. The dashed line marks the mid-run proposer rewrite.}
  \label{fig:translate_loss_refusal}
\end{figure}

\begin{figure}[htbp]
  \centering
  \begin{subfigure}{0.48\linewidth}\includegraphics[width=\linewidth]{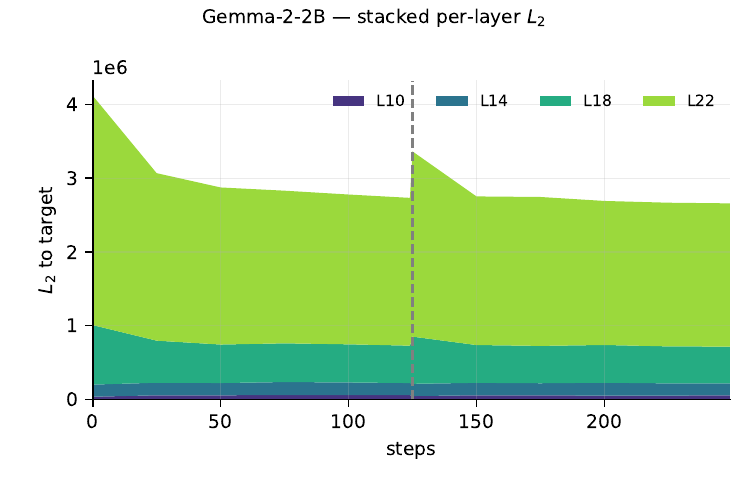}\end{subfigure}\hfill
  \begin{subfigure}{0.48\linewidth}\includegraphics[width=\linewidth]{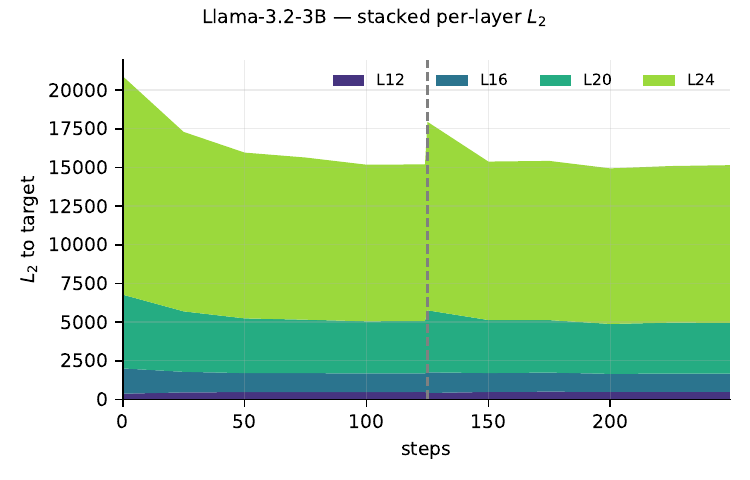}\end{subfigure}
  \caption{Stacked per-layer $L_2$ distance to the target for the \emph{model-translation}
           anti-refusal run.}
  \label{fig:mt_stack_refusal}
\end{figure}

\begin{figure}[htbp]
  \centering
  \begin{subfigure}{0.48\linewidth}\includegraphics[width=\linewidth]{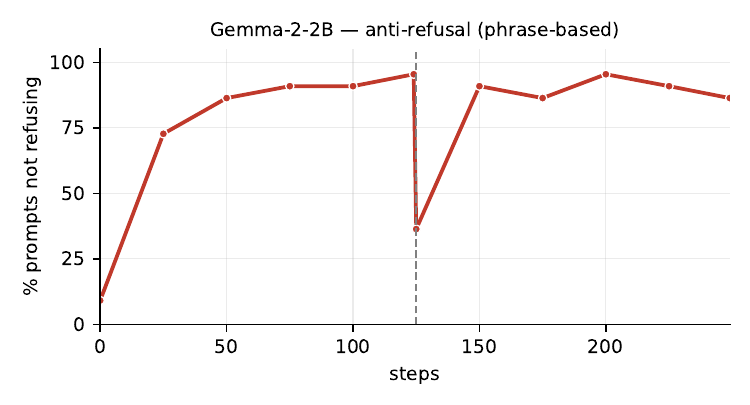}\end{subfigure}\hfill
  \begin{subfigure}{0.48\linewidth}\includegraphics[width=\linewidth]{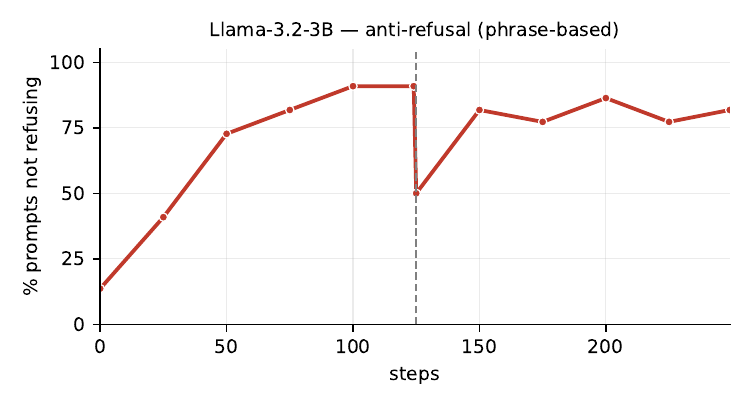}\end{subfigure}
  \caption{Fraction of train prompts whose output is not a phrase-based refusal every $25$ steps for the model-translation run.}
  \label{fig:mt_switch_refusal}
\end{figure}

\FloatBarrier
\paragraph{Model translation.} We then track the change of loss for our model-translation
experiments. Figure~\ref{fig:translate_loss_refusal} sums the $L_2$ distance and averages the
cosine over the layers, Figure~\ref{fig:mt_stack_refusal} breaks the $L_2$ down per layer, and
Figure~\ref{fig:mt_switch_refusal} tracks the change in refusal rate. The change of suffix in the
middle of the steps by the proposer sharply increases the loss. Despite the fact that the safety
metric performs significantly worse than its German counterpart, we do not notice a strong
difference between the losses for German and for refusal. This may be an indication that the
steering vector for refusal is more related to avoiding the phrase substrings than pure
jailbreaking.

\section{Soft suffixes}
\label{sec:models-judging} 

To check how removing the discrete token constraint performs, we replace the $20$ discrete suffix tokens with $20$ \emph{soft} tokens, and optimise them directly with Adam
(with learning rate $0.05$ and $150$ steps) against the same activation target as ActGCG (Equation~\ref{eq:ares-loss}). We see this as an indication on how well the objective can be met, since the optimiser explores the full continuous embedding space. We run this on Gemma-2-2B and Qwen3-1.7B, and the same $30$ German prompts, at both $\alpha = 1.0$ and the $\alpha = 2.0$.
 
 \begin{table}[htbp]
    \centering
    \caption{Soft suffix German success rate at two steering strengths ($30$ prompts).}
    \label{tab:soft-german}
    \begin{tabular}{l r r}
      \toprule
      Model & {$\alpha = 1.0$ \pct} & {$\alpha = 2.0$ \pct} \\
      \midrule
      Gemma-2-2B  & 96.67 & 93.33 \\
      Qwen3-1.7B  & 93.33 & 70.00 \\
      \bottomrule
    \end{tabular}
  \end{table}

At $\alpha = 1.0$, the soft suffix reaches near perfect German rates compared to its discrete version with ActGCG. This shows that the discrete search over tokens is a limitation. For $\alpha = 2.0$, we note a peculiarity where the success rate of Qwen3-1.7B decreases strongly due to empty outputs, likely due to the vector being too strong, driving the activations to extreme values which the model cannot decode.  

\section{Decoding soft tokens}

We tried to convert the optimised soft embeddings back into discrete tokens by projecting each soft token to its nearest vocabulary embedding (using cosine similarity) and evaluating the resulting discrete string as a suffix. This attempt failed completely, as the suffix produced zero German responses. Hence, a soft suffix cannot be easily converted into a discrete one, which justifies why a discrete search is useful to obtain a hard suffix.

\section{Different dataset extraction for German}

We change how the German steering vector is extracted. The default construction (the \emph{German only} column) contrasts a German prompt with a German answer against an English prompt with an English answer. We change the way we extract the dataset such that we hold the prompt fixed in English and contrast only the answer language (the \emph{English + German} column), with the positive example being an English prompt followed by its German answer, and the negative example being the same English prompt followed by its original English answer (\texttt{en\_prompt + de\_answer} against \texttt{en\_prompt + en\_answer}). The vector is built from the same Dolly dataset and at the same four layers as before, and we then run per-prompt ActGCG on Gemma-2-2B and Qwen3-1.7B for $30$ prompts. We use $\alpha = 1.0$ rather than the German default of $2.0$. 

\begin{table}[htbp]
  \centering
  \caption{Per-prompt German success rate with $\alpha = 1.0$ and $30$ prompts, comparing two different Dolly dataset extraction methods.}
  \label{tab:en-ger}
  \begin{tabular}{l r r}
    \toprule
    Model & {German only \pct} & {English + German \pct} \\
    \midrule
    Qwen3-1.7B   & 80.00 & 83.33 \\
    Gemma-2-2B & 53.33 & 53.33 \\
    \bottomrule
  \end{tabular}
\end{table}

The new dataset variant shows no consistent or clear improvement. Since we started with the German only method, we simply stay with the original way we extract the Dolly dataset.

\section{More steps}

We test whether the default $250$ optimisation steps are enough for refusal, or whether the discrete search continues to improve, by re-running the per-prompt anti-refusal experiment at $500$ steps with $\alpha = 1.0$, and breadth split using $22$ prompts.
  
  \begin{table}[htbp] 
    \centering
    \caption{Per-prompt anti-refusal, breadth split. HarmBench jailbreak rate \pct\ at $250$ against $500$ optimisation steps.}
    \label{tab:more-steps}
    \begin{tabular}{l r r}
      \toprule
      Model & $250$ steps & $500$ steps \\
      \midrule
      Gemma-2-2B  & 27.3 & 31.8 \\
      Qwen3-1.7B  & 40.9 & 54.5 \\
      \bottomrule
    \end{tabular}
  \end{table}

More steps do increase the success rate. Nevertheless, computation time increases significantly as well. Hence, we continue to use $250$ steps throughout our experiments.

\section{Weight precision}

We test whether the model's weight precision, the number of bits per weight, affects our method's search. We run the per-prompt German experiment on Gemma-2-2B in full \texttt{fp32} precision instead of the default \texttt{bf16}, keeping everything else fixed while using $30$ prompts. Note that this run uses $\alpha = 1.0$ rather than the $\alpha = 2.0$ of the main German results, but we mainly consider the difference between the two precisions at the same $\alpha$.

\begin{table}[htbp]
  \centering
  \caption{Effect of weight precision at $\alpha = 1.0$ (Gemma-2-2B).}
  \label{tab:precision}
  \begin{tabular}{l r}
    \toprule
    Precision & {Success \pct} \\
    \midrule
    \texttt{bf16} (default) & 53.33 \\
    \texttt{fp32}           & 70.00 \\
    \bottomrule
  \end{tabular}
\end{table}

We notice that \texttt{fp32} precision does perform better. However, since it takes a significantly longer amount of computation time, and the gap is within the seed-to-seed variance seen earlier, we continue to use \texttt{bf16} throughout our experiments.

\section{Suffix to steering vectors}

To complete the equivalence relationship in the reverse direction, we ask whether a suffix's effect can itself be distilled back into a steering vector. For each of the $30$ Gemma-2-2B German prompts, we take its optimised per-prompt ActGCG suffix and measure the activation difference it induces at the filler positions of the four target layers, where we take the activations with the suffix appended minus those without. We obtain a \emph{recovered} steering vector, and add it at the filler positions during generation exactly as we would a normal steering vector, and measure the success rate. Here, we compare two variants: a \emph{per-prompt} recovered vector (each prompt steered by the vector recovered from its own suffix) and the \emph{mean} recovered vector (a single vector averaged over all $30$ prompts, applied to every prompt), each for $\alpha = 1.0$ and $\alpha = 2.0$.

\begin{table}[htbp]
  \centering
  \caption{Success rate of steering vectors recovered from a suffix (Gemma-2-2B, German, $30$ prompts).}
  \label{tab:suffix-to-sv}
  \begin{tabular}{r r r}
    \toprule
    {$\alpha$} & {Mean SV} & {Per-prompt SV} \\
               & {\pct} & {\pct} \\
    \midrule
    1.0 & 66.67 & 50.00 \\
    2.0 & 80.00 & 63.33 \\
    \bottomrule
  \end{tabular}
\end{table}

A steering vector recovered from the suffix does reproduce the behaviour, and the mean vector recovers it better than the per-prompt vector. Although we do observe a slight decrease in effectiveness, this supports the equivalence between the suffix and the steering vector.

\section{Prefix instead of suffix}

We check whether optimising a \emph{prefix} placed before a prompt performs better than a \emph{suffix} placed after the prompt. Unlike a suffix, a prefix can, in principle, match an activation target that also covers the prompt positions. Therefore, we compare two targets for the prefix: the steering vector added only at the filler positions, and the steering vector added at both the prompt and filler positions. We evaluate on German with Gemma-2-2B for $30$ prompts with $\alpha \in \{1.0, 2.0, 3.0\}$ 

\begin{table}[htbp]
  \centering
  \caption{Prefix optimisation on German with Gemma-2-2B for a filler only target against a prompt and filler target.}
  \label{tab:filler-vs-prompt}
  \begin{tabular}{l r r r}
    \toprule
    Target & {$\alpha = 1.0$} & {$\alpha = 2.0$} & {$\alpha = 3.0$} \\
           & {\pct} & {\pct} & {\pct} \\
    \midrule
    Filler only     & 16.67 & 60.00 & 66.67 \\
    Prompt and filler & 23.33 & 33.33 & 30.00 \\
    \bottomrule
  \end{tabular}
\end{table}

Even though a prefix can reach the prompt positions, matching the prompt and filler target is generally worse than matching the filler positions alone. Moreover, the best prefix result is still behind its suffix counterpart. Therefore, the suffix remains the stronger choice.
  
\section{Starting from a readable German instruction}

Until now, we have initialised the optimised suffix from the string \texttt{x x \ldots x}. We check whether starting from a \emph{readable} instruction that already asks for the target behaviour gives ActGCG an easier starting point and a higher final success rate. We run the per-prompt German experiment for the four main models with only the initial suffix changed, keeping everything else fixed ($30$ Dolly prompts, $\alpha = 2.0$, seed 0). The initial suffix is ``Please answer in German''.

\begin{table}[htbp]
  \centering
  \caption{Per-prompt German success rate \pct\ when the suffix is initialised from \texttt{x x \ldots x} against a readable instruction.}
  \label{tab:readable-init}
  \begin{tabular}{l r r}
    \toprule
    Model & \texttt{x x \ldots x} & \texttt{Please answer in German} \\
    \midrule
    Gemma-2-2B    & 86.67  & 86.67 \\
    Llama-3.2-3B  & 100.00 & 100.00 \\
    Qwen3-1.7B    & 96.67  & 90.00 \\
    Qwen3-4B      & 96.67  & 93.33 \\
    \bottomrule
  \end{tabular}
\end{table}

A target-oriented initialisation makes essentially no difference. This shows that GCG, as well as our ActGCG method, does not preserve coherence between suffix tokens, as it only optimises them independently.

\chapter{Discussion}%
\label{chap:discussion}
\section{Limitations}

Our results come with several limitations. The most fundamental is that the success rate depends heavily on the \emph{concept} being steered, and we test only two of them here: the German language and refusal. These two already behave very differently: German is reproduced almost perfectly by a suffix and transfers well across prompts and models, whereas refusal is weaker. From here, it is not possible to infer how other concepts would perform.

The evaluations are also run at a relatively small scale. The per-prompt and universal experiments were tested on $30$ prompts for German, and between $22$ and $60$ for refusal. Most runs use a single fixed seed. As a result, our scores may vary with different prompts or seeds.

By construction, ActGCG searches over discrete tokens, which limits how closely it can match the target activations. Our soft suffix experiments, which optimise continuous embeddings directly, reach lower losses than any discrete suffix and act as an upper bound on what the objective can achieve. The gap between them reflects the reduction in success rate due to the discrete token space.

The method is also bounded by the steering vector it imitates. The suffix can only reproduce a behaviour that the steering vector itself induces. If the vector is weak, the suffix inherits that ceiling as well. Moreover, reaching the activation target is not sufficient to guarantee the behaviour. For instance, a suffix can match the target activations closely and still fail to jailbreak. Matching the activations is thus a useful proxy for the behaviour, but not a guarantee of it. Furthermore, the best way to construct a steering vector in the first place is still an open question. A more optimal steering vector could further improve our method.

Finally, judging the responses is not perfect. Especially for refusal, the measured safety rate depends heavily on the judge. We standardise on a single classifier throughout our experiments to keep the runs comparable. The phrase based metric is not a good proxy for safety either, as a response can often avoid refusal phrases without delivering harmful content. For German, the \texttt{langdetect} judge occasionally mislabels short responses that are German overall but contain English titles or names. In addition, the definition of the concept itself matters, and results can vary drastically across the literature due to the difficulty of rigorously defining behaviours.

\section{Ethical Considerations}

Our work considers the anti-refusal direction, which can sometimes result in bypassing safety guardrails. We conduct all of our jailbreak experiments on small, open-weight models that are already publicly available, and we do not present any harmful model outputs in this thesis. The practical offensive value of the resulting suffixes is limited. They jailbreak only at modest rates and, crucially, transfer poorly across models. Therefore, a suffix optimised on one model is of little use against another.

Our aim in understanding an equivalence relationship between suffixes and steering vectors is defensive. Showing that a suffix can imitate a steering vector is a step towards building measures against inputs that reproduce vectors of malicious concepts.

\section{Future Work}

Our findings suggest a number of directions worth pursuing. The most direct option would be to test more concepts, and to check how well our approach generalises. German reaches a high success rate while refusal shows highly mixed results; hence, it would be informative to see how other concepts fall on this scale.

The matching objective could also be further improved. We use a squared-$L_2$ loss, but there may be other losses that can approximate the target activation more closely. Since squaring amplifies large deviations, a more robust objective could reduce their influence.

Furthermore, the steering vector itself could be strengthened by searching for better construction methods than the difference-of-means vectors. More
broadly, steering vectors themselves may not be the best proxy for a given behaviour. Our method draws its target activation from a steering vector, but for instance, we could also choose activations from another model that already exhibits the desired behaviour. Investigating such alternative targets with input suffixes would test whether there are equivalence relationships for behaviours that a steering vector does not fully capture.

\chapter{Conclusion}%
\label{chap:conclusion}

In this thesis, we investigate whether there is an equivalence between input suffixes and steering vectors. We introduce ActGCG, a modification of GCG that, rather than optimising a suffix towards a fixed output string, optimises it such that its activations match the target obtained by adding a steering vector to the residual stream. We test this hypothesis on two concepts, namely the German language and refusal. For German, we largely retain the behaviour of the steering vector with a discrete suffix using our method. For refusal, the equivalence is weaker. Our method largely suppresses the model's refusal, such that it succeeds under a substring-based metric that only checks whether the response avoids a set of refusal phrases. However, when measured with the stricter HarmBench classifier, the suffix that reproduces the steering vector does not directly produce a jailbreak. The substring metric roughly doubles the true HarmBench rate.

We find the same asymmetry between the two concepts throughout the rest of our results. We extend the method to universal suffixes that steer many prompts at once, and to a single suffix shared across models, and find that our behaviour is still retained. The German suffix transfers well both to test prompts and to other models, whereas the anti-refusal suffix generalises across prompts but transfers poorly across models. The same pattern appears when we ask a larger model to rewrite the unreadable suffix into coherent text midway through optimisation. The model translation improves success for German, most significantly for transferring to different models, but gives only mixed results for refusal. All in all, our results show that a steering vector's behaviour is reachable from the input, with no access to the model's internals at generation time, but with a certain loss depending on the concept. Establishing how far this equivalence extends to further concepts beyond the two studied here, to better steering vector constructions, as well as to alternative losses, remains a promising direction for the future.

\bibliographystyle{plainnat}
\bibliography{Content/Bibliography}

@inproceedings{mikolov2013b,
  title     = {Distributed Representations of Words and Phrases and their Compositionality},
  author    = {Mikolov, Tom{\'a}s and Sutskever, Ilya and Chen, Kai and Corrado, Greg S. and Dean, Jeffrey},
  booktitle = {Advances in Neural Information Processing Systems 26 (NeurIPS)},
  pages     = {3111--3119},
  year      = {2013},
  publisher = {Curran Associates, Inc.}
}

@inproceedings{pennington2014,
  title     = {GloVe: Global Vectors for Word Representation},
  author    = {Pennington, Jeffrey and Socher, Richard and Manning, Christopher D.},
  booktitle = {Proceedings of the 2014 Conference on Empirical Methods in Natural Language Processing (EMNLP)},
  pages     = {1532--1543},
  year      = {2014},
  address   = {Doha, Qatar},
  publisher = {Association for Computational Linguistics}
}

@inproceedings{bolukbasi2016,
  title     = {Man is to Computer Programmer as Woman is to Homemaker? Debiasing Word Embeddings},
  author    = {Bolukbasi, Tolga and Chang, Kai-Wei and Zou, James and Saligrama, Venkatesh and Kalai, Adam},
  booktitle = {Advances in Neural Information Processing Systems 29 (NeurIPS)},
  pages     = {4349--4357},
  year      = {2016},
  publisher = {Curran Associates, Inc.}
}

@article{elhage2022,
  title     = {Toy Models of Superposition},
  author    = {Elhage, Nelson and Hume, Tristan and Olsson, Catherine and Schiefer, Nicholas and Henighan, Tom and others},
  journal   = {Transformer Circuits Thread},
  year      = {2022},
  publisher = {Anthropic}
}

@inproceedings{park2023,
  title   = {The Linear Representation Hypothesis and the Geometry of Large Language Models},
  author  = {Park, Kiho and Choe, Yo Joong and Veitch, Victor},
  booktitle = {Proceedings of the 41st International Conference on Machine Learning (ICML)},
  year    = {2024}
}

@inproceedings{nanda2023,
  title     = {Emergent Linear Representations in World Models of Self-Supervised Sequence Models},
  author    = {Nanda, Neel and Lee, Andrew and Wattenberg, Martin},
  booktitle = {Proceedings of the 6th BlackboxNLP Workshop: Analyzing and Interpreting Neural Networks for NLP},
  pages     = {16--30},
  year      = {2023},
  address   = {Singapore},
  publisher = {Association for Computational Linguistics}
}

@article{zou2023gcg,
  title   = {Universal and Transferable Adversarial Attacks on Aligned Language Models},
  author  = {Zou, Andy and Wang, Zifan and Carlini, Nicholas and Nasr, Milad and Kolter, J. Zico and Fredrikson, Matt},
  journal = {arXiv preprint arXiv:2307.15043},
  year    = {2023}
}

@inproceedings{rimsky2024caa,
  title     = {Steering Llama 2 via Contrastive Activation Addition},
  author    = {Rimsky, Nina and Gabrieli, Nick and Schulz, Julian and Tong, Meg and Hubinger, Evan and Turner, Alexander Matt},
  booktitle = {Proceedings of the 62nd Annual Meeting of the Association for Computational Linguistics (ACL)},
  year      = {2024}
}

@article{mazeika2024harmbench,
  title   = {HarmBench: A Standardized Evaluation Framework for Automated Red Teaming and Robust Refusal},
  author  = {Mazeika, Mantas and Phan, Long and Yin, Xuwang and Zou, Andy and Wang, Zifan and Mu, Norman and Sakhaee, Elham and Li, Nathaniel and Basart, Steven and Li, Bo and Forsyth, David and Hendrycks, Dan},
  journal = {Proceedings of the 41st International Conference on Machine Learning (ICML)},
  year    = {2024}
}

@misc{databricks2023dolly,
  title        = {Free Dolly: Introducing the World's First Truly Open Instruction-Tuned LLM},
  author       = {{Databricks}},
  year         = {2023},
  howpublished = {\url{https://www.databricks.com/blog/2023/04/12/dolly-first-open-commercially-viable-instruction-tuned-llm}}
}

@inproceedings{vaswani2017attention,
  title     = {Attention Is All You Need},
  author    = {Vaswani, Ashish and Shazeer, Noam and Parmar, Niki and Uszkoreit, Jakob and Jones, Llion and Gomez, Aidan N. and Kaiser, {\L}ukasz and Polosukhin, Illia},
  booktitle = {Advances in Neural Information Processing Systems 30 (NeurIPS)},
  pages     = {5998--6008},
  year      = {2017},
  publisher = {Curran Associates, Inc.}
}

@article{elhage2021framework,
  title     = {A Mathematical Framework for Transformer Circuits},
  author    = {Elhage, Nelson and Nanda, Neel and Olsson, Catherine and Henighan, Tom and Joseph, Nicholas and Mann, Ben and Askell, Amanda and Bai, Yuntao and Chen, Anna and Conerly, Tom and DasSarma, Nova and Drain, Dawn and Ganguli, Deep and Hatfield-Dodds, Zac and Hernandez, Danny and Jones, Andy and Kernion, Jackson and Lovitt, Liane and Ndousse, Kamal and Amodei, Dario and Brown, Tom and Clark, Jack and Kaplan, Jared and McCandlish, Sam and Olah, Chris},
  journal   = {Transformer Circuits Thread},
  year      = {2021},
  publisher = {Anthropic}
}

@article{im2025unified,
  title   = {A Unified Understanding and Evaluation of Steering Methods},
  author  = {Im, Shawn and Li, Sharon},
  journal = {arXiv preprint arXiv:2502.02716},
  year    = {2025}
}

@article{pres2024reliable,
  title   = {Towards Reliable Evaluation of Behavior Steering Interventions in LLMs},
  author  = {Pres, Itamar and Ruis, Laura and Lubana, Ekdeep Singh and Krueger, David},
  journal = {arXiv preprint arXiv:2410.17245},
  year    = {2024}
}

@article{li2024translation,
  title   = {Deciphering the Chaos: Enhancing Jailbreak Attacks via Adversarial Prompt Translation},
  author  = {Li, Qizhang and Yang, Xiaochen and Zuo, Wangmeng and Guo, Yiwen},
  journal = {arXiv preprint arXiv:2410.11317},
  year    = {2024}
}

@article{fiotto-kaufman2024nnsight,
  title   = {NNsight and NDIF: Democratizing Access to Open-Weight Foundation Model Internals},
  author  = {Fiotto-Kaufman, Jaden and Loftus, Alexander R. and Todd, Eric and Brinkmann, Jannik and Pal, Koyena and Troitskii, Dmitrii and Ripa, Michael and others},
  journal = {arXiv preprint arXiv:2407.14561},
  year    = {2024}
}

@article{touvron2023llama2,
  title   = {Llama 2: Open Foundation and Fine-Tuned Chat Models},
  author  = {Touvron, Hugo and Martin, Louis and Stone, Kevin and Albert, Peter and Almahairi, Amjad and Babaei, Yasmine and Bashlykov, Nikolay and Batra, Soumya and Bhargava, Prajjwal and Bhosale, Shruti and others},
  journal = {arXiv preprint arXiv:2307.09288},
  year    = {2023}
}

@inproceedings{arditi2024refusal,
  title   = {Refusal in Language Models Is Mediated by a Single Direction},
  author  = {Arditi, Andy and Obeso, Oscar and Syed, Aaquib and Paleka, Daniel and Panickssery, Nina and Gurnee, Wes and Nanda, Neel},
  booktitle = {Advances in Neural Information Processing Systems 37 (NeurIPS)},
  year    = {2024}
}

@article{turner2023actadd,
  title   = {Steering Language Models With Activation Engineering},
  author  = {Turner, Alexander Matt and Thiergart, Lisa and Leech, Gavin and Udell, David and Vazquez, Juan J. and Mini, Ulisse and MacDiarmid, Monte},
  journal = {arXiv preprint arXiv:2308.10248},
  year    = {2023}
}

@inproceedings{li2023iti,
  title     = {Inference-Time Intervention: Eliciting Truthful Answers from a Language Model},
  author    = {Li, Kenneth and Patel, Oam and Vi{\'e}gas, Fernanda and Pfister, Hanspeter and Wattenberg, Martin},
  booktitle = {Advances in Neural Information Processing Systems 36 (NeurIPS)},
  year      = {2023}
}

@article{zou2023repe,
  title   = {Representation Engineering: A Top-Down Approach to AI Transparency},
  author  = {Zou, Andy and Phan, Long and Chen, Sarah and Campbell, James and Guo, Phillip and Ren, Richard and Pan, Alexander and Yin, Xuwang and Mazeika, Mantas and others},
  journal = {arXiv preprint arXiv:2310.01405},
  year    = {2023}
}

@article{winninger2025ssr,
  title   = {Using Mechanistic Interpretability to Craft Adversarial Attacks against Large Language Models},
  author  = {Winninger, Thomas and Addad, Boussad and Kapusta, Katarzyna},
  journal = {arXiv preprint arXiv:2503.06269},
  year    = {2025}
}

@inproceedings{wen2023pez,
  title   = {Hard Prompts Made Easy: Gradient-Based Discrete Optimization for Prompt Tuning and Discovery},
  author  = {Wen, Yuxin and Jain, Neel and Kirchenbauer, John and Goldblum, Micah and Geiping, Jonas and Goldstein, Tom},
  booktitle = {Advances in Neural Information Processing Systems 36 (NeurIPS)},
  year    = {2023}
}

@article{gemmateam2024gemma2,
  title   = {Gemma 2: Improving Open Language Models at a Practical Size},
  author  = {{Gemma Team}},
  journal = {arXiv preprint arXiv:2408.00118},
  year    = {2024}
}

@article{yang2025qwen3,
  title   = {Qwen3 Technical Report},
  author  = {Yang, An and others},
  journal = {arXiv preprint arXiv:2505.09388},
  year    = {2025}
}

@article{qwen2024qwen25,
  title   = {Qwen2.5 Technical Report},
  author  = {Yang, An and others},
  journal = {arXiv preprint arXiv:2412.15115},
  year    = {2024}
}

@article{grattafiori2024llama3,
  title   = {The Llama 3 Herd of Models},
  author  = {Grattafiori, Aaron and others},
  journal = {arXiv preprint arXiv:2407.21783},
  year    = {2024}
}

@inproceedings{kingma2015adam,
  title     = {Adam: A Method for Stochastic Optimization},
  author    = {Kingma, Diederik P. and Ba, Jimmy},
  booktitle = {International Conference on Learning Representations (ICLR)},
  year      = {2015}
}

@misc{grayswan2024nanogcg,
    title        = {{nanoGCG}: A fast and lightweight implementation of the {GCG} algorithm in {PyTorch}},
    author       = {{Gray Swan AI}},
    year         = {2024},
    howpublished = {\url{https://github.com/GraySwanAI/nanoGCG}},
    note         = {Accessed: 2026-08-25}
}

@misc{danilak2014langdetect,
    title        = {langdetect: Port of Google's language-detection library to {Python}},
    author       = {Danilak, Michal},
    year         = {2014},
    howpublished = {\url{https://github.com/Mimino666/langdetect}}, 
    note         = {Accessed: 2026-08-25}
}

\appendix
\chapter{Additional Figures}
\section{Per-layer cosine trajectories}\label{app:perlayer-cosine}
\begin{figure}[htbp]
  \centering
  \includegraphics[width=0.9\linewidth]{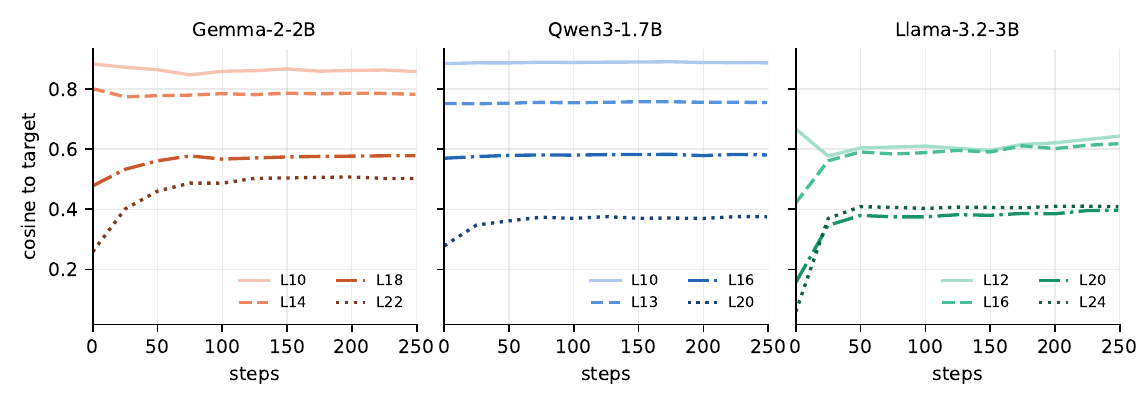}
  \caption{Per-layer cosine to the target activations over optimisation steps for the universal German run.}
  \label{fig:uni_loss_per_layer}
\end{figure}

\begin{figure}[htbp] 
    \centering
    \includegraphics[width=0.9\linewidth]{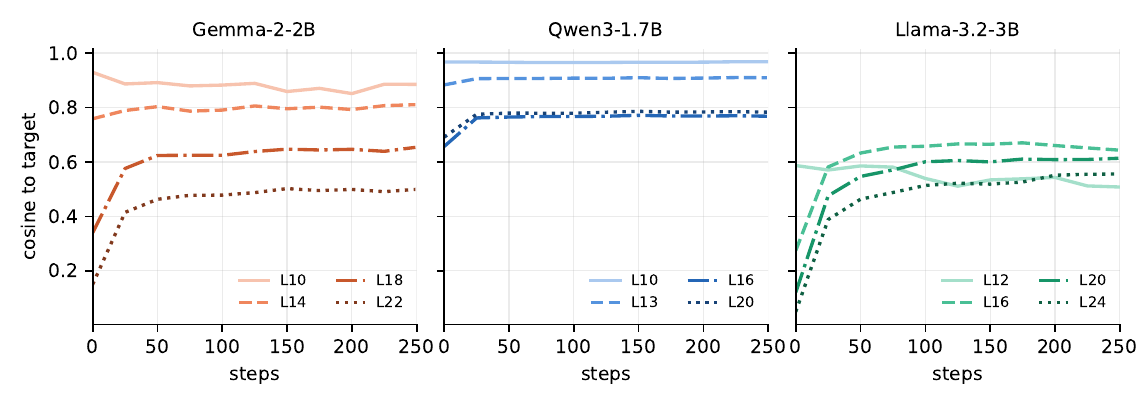}
    \caption{Per-layer cosine to the target activations over optimisation steps for the universal anti-refusal run.}
    \label{fig:uni_loss_refusal_per_layer}
  \end{figure}

\section{Per-prompt loss trajectories}\label{app:loss-grid}

\begin{figure}[htbp]
    \centering
    \includegraphics[width=\linewidth]{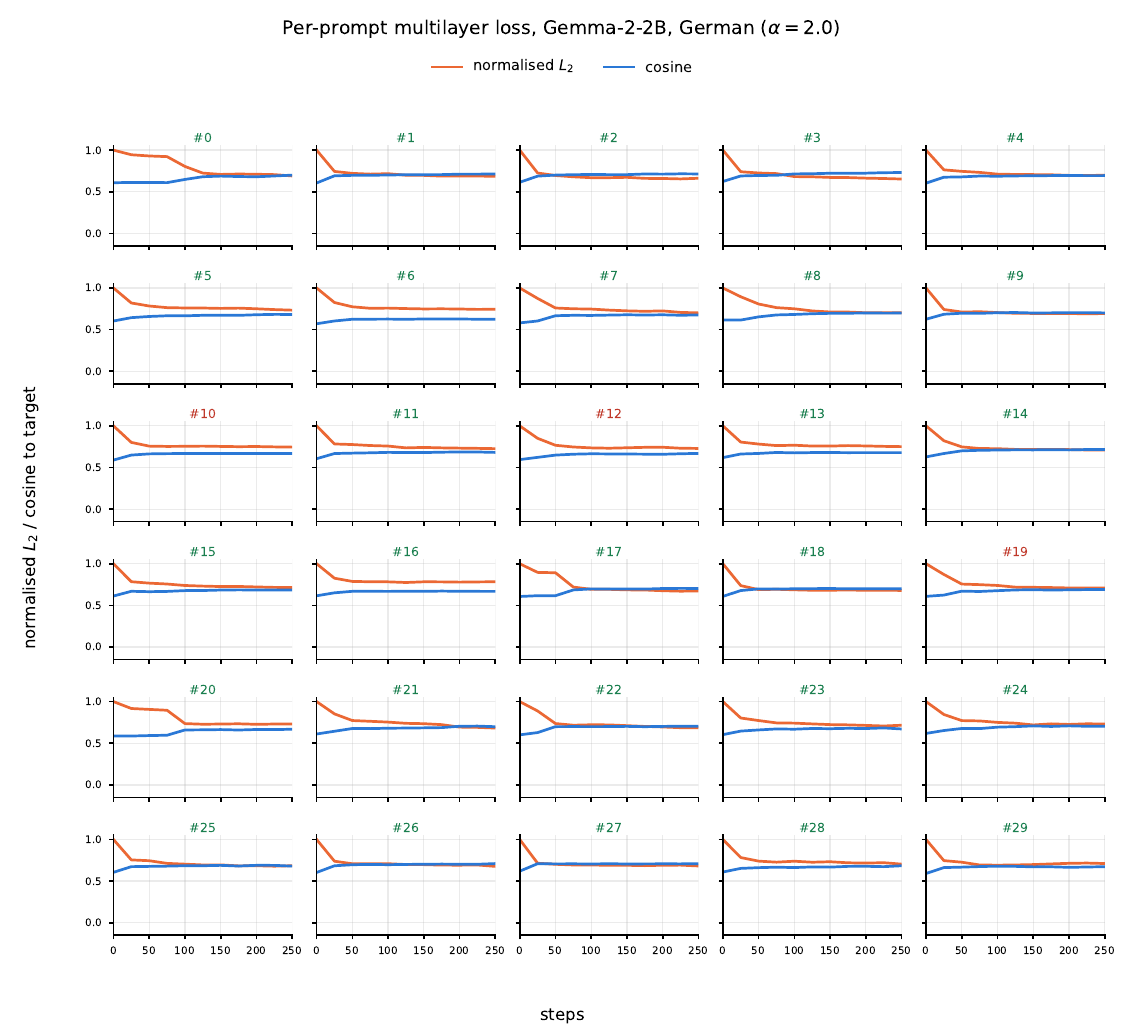}
    \caption{Per-prompt multilayer loss for German runs (Gemma-2-2B, $\alpha = 2.0$,
             $30$ prompts), with normalised $L_2$ and cosine to the target over optimisation steps. Panel titles are green where the answer is German, red otherwise.}
    \label{fig:loss-grid-german}
  \end{figure}
  
  \begin{figure}[htbp]
    \centering
    \includegraphics[width=\linewidth]{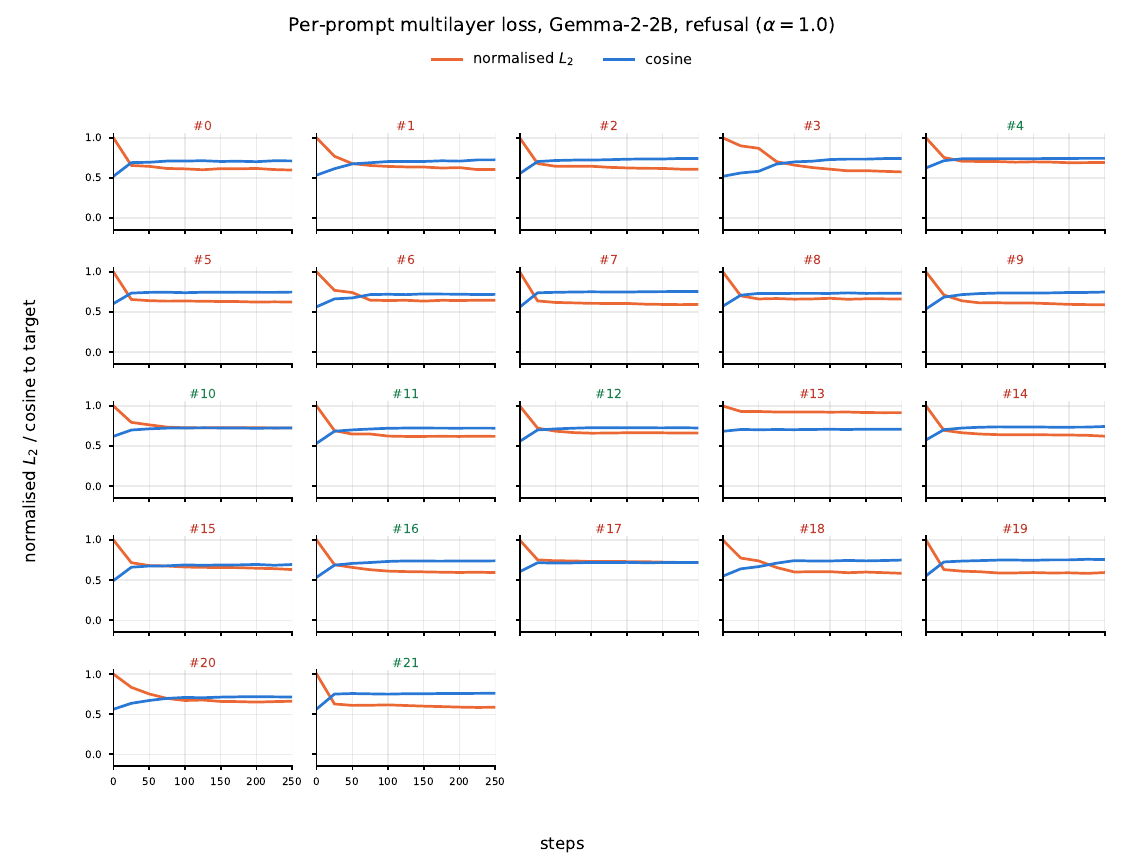}
    \caption{Per-prompt multilayer loss for the anti-refusal runs (Gemma-2-2B,
             $\alpha = 1.0$, $22$ prompts), with normalised $L_2$ and cosine to the target over optimisation steps. Panel titles are green where the prompt is a jailbreak judged by HarmBench, red otherwise.}
    \label{fig:loss-grid-refusal}
\end{figure}

\end{document}